\documentclass{article}

\usepackage[final]{corl_2026} 
\usepackage{graphicx}

\usepackage{makecell} 
\IfFileExists{ifsym.sty}{\usepackage[misc]{ifsym}}{\usepackage{marvosym}}     

\usepackage{booktabs}                                   
\usepackage{multirow}                                   
\usepackage{tablefootnote}                              
\usepackage[symbol]{footmisc}                           
\usepackage{amsmath,amssymb}                            
\usepackage{xcolor}                                     

\usepackage{enumitem}                                   
\usepackage{subcaption}                                 
\usepackage{caption}     

\usepackage{graphicx}

\usepackage{cleveref}

\usepackage{csquotes}

\usepackage{tikz}
\usetikzlibrary{shapes, arrows.meta, positioning}
\usepackage{wrapfig}

\newcommand{\eat}[1]{}                                  

\usepackage{nicematrix}
\usepackage{bm}
\usepackage{pifont}

\usepackage{hyperref}

\newtheorem{theorem}{Theorem}[section]

\newcommand{\bcode}      {e}

\title{RoboAug: One Annotation to Hundreds of Scenes via Region-Contrastive Data Augmentation for Robotic Manipulation}

\author{%
  \parbox{\textwidth}{%
    \centering
    \normalfont\small
    Xinhua Wang\textsuperscript{1,*},
    Kun Wu\textsuperscript{1,*},
    Zhen Zhao\textsuperscript{1,*},
    Hu Cao\textsuperscript{2},
    Yinuo Zhao\textsuperscript{1,3},
    Zhiyuan Xu\textsuperscript{1},
    Meng Li\textsuperscript{1},\\[2pt]
    Shichao Fan\textsuperscript{1,4},
    Di Wu\textsuperscript{1,5},
    Yixue Zhang\textsuperscript{1,6},
    Ning Liu\textsuperscript{1},
    Zhengping Che\textsuperscript{1,\dag,\Letter},
    Jian Tang\textsuperscript{1,\Letter}
    \\[5pt]
    \footnotesize
    \mbox{\textsuperscript{1}Beijing Innovation Center of Humanoid Robotics}\\
    \mbox{\textsuperscript{2}School of Automation, Southeast University}\quad
    \mbox{\textsuperscript{3}City University of Hong Kong}\\
    \mbox{\textsuperscript{4}The School of Mechanical Engineering and Automation, Beihang University}\\
    \mbox{\textsuperscript{5}State Key Laboratory of Multimedia Information Processing,
    School of Computer Science, Peking University}\\
    \mbox{\textsuperscript{6}The School of Advanced Manufacturing and Robotics, Peking University}
    \\[3pt]
    \textsuperscript{*}Equal contribution
    \quad
    \textsuperscript{\dag}Team leader
    \quad
    \textsuperscript{\Letter}Corresponding authors
    \\[3pt]
    \texttt{\{will.wang, z.che\}}@x-humanoid.com}
  }%

\begin{document}

\makeatletter
\let\@oldmaketitle\@maketitle%
\renewcommand{\@maketitle}{\@oldmaketitle%
    \centering
    \vspace{-2.5em} 
    \includegraphics[width=0.95\linewidth]{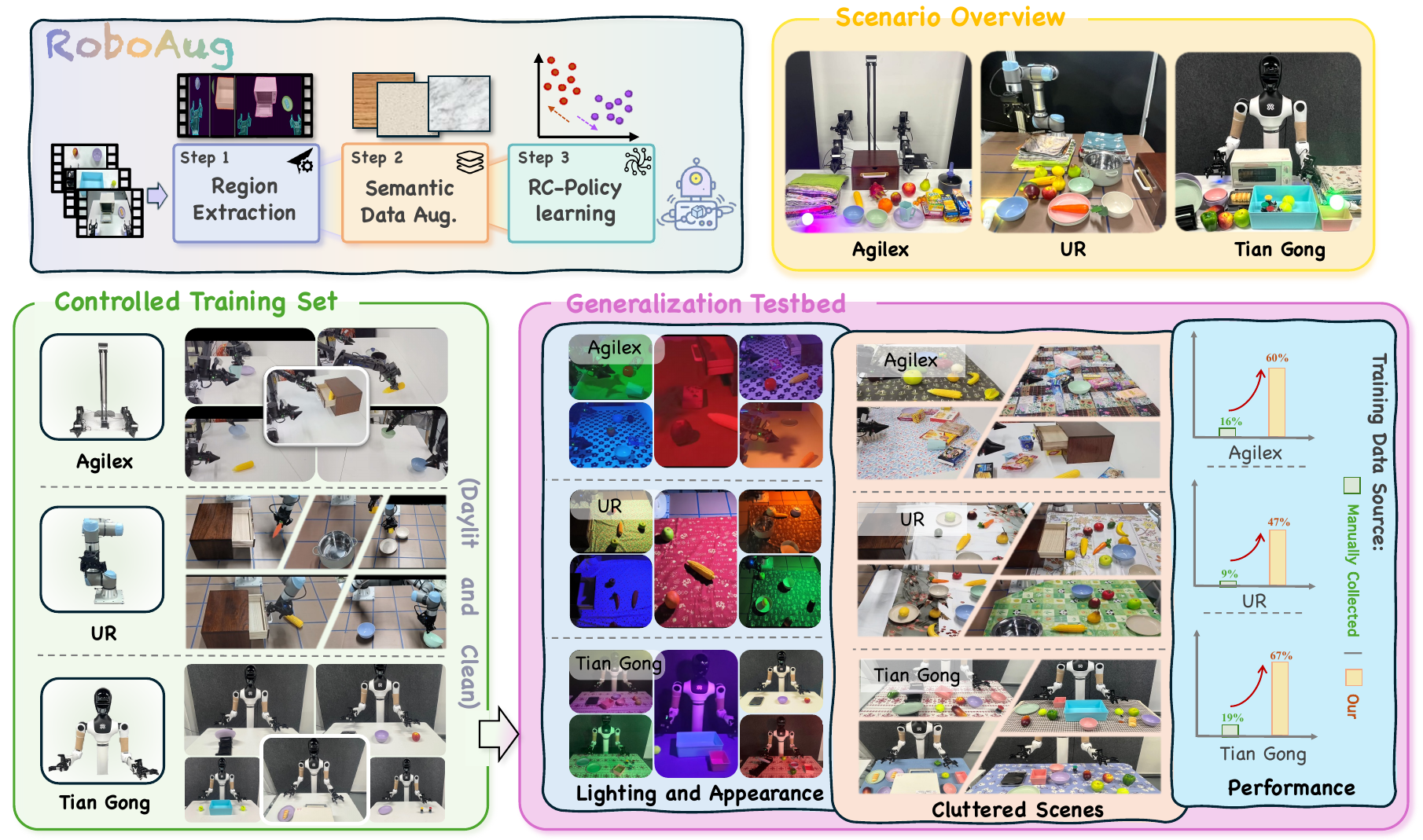}
    \captionof{figure}{We introduce RoboAug, a region-contrastive data augmentation framework. RoboAug enables robust robotic generalization in diverse, unseen scenes.}
    \label{fig:abs}
}
\makeatother

\maketitle


\begin{abstract}
Enhancing the generalization of robotic learning in diverse unseen environments remains a fundamental challenge. Existing approaches often rely on large-scale pretraining, which is labor-intensive and time-consuming, or semantic data augmentation methods that assume flawless upstream object detection in real-world scenarios. In this work, we propose RoboAug, a novel generative data augmentation framework that reduces reliance on large-scale pretraining and perfect visual recognition by requiring only a single image with bounding box annotations for dataset construction. Leveraging this minimal supervision, RoboAug employs pretrained generative models for precise semantic augmentation and introduces a plug-and-play region-contrastive loss to guide attention toward task-relevant regions, thereby enhancing generalization and task success rates. Extensive real-world experiments on UR-5e, AgileX, and Tian Gong 2.0 demonstrate that RoboAug consistently outperforms state-of-the-art augmentation baselines under background, distractor, and lighting shifts. 
Our project is available at https://x-roboaug.github.io/.

\end{abstract}



\section{Introduction}
\vspace{-5pt}
Generalist robots operating in unstructured real-world environments require perceptual robustness beyond controlled training conditions. Despite strong performance on complex skills~\cite{Brohan-RSS-23,zitkovich2023rt,zhao2023learning,o2024open,rss2025_avisionlanguagea, fan2025xr}, end-to-end visuomotor policies remain vulnerable to observational distribution shifts, often suffering severe degradation in unseen environments. To address this issue, we improve policy generalization against three major sources of out-of-distribution (OOD) interference: \textit{background variations}, \textit{lighting changes}, and \textit{task-irrelevant distractors}.


To tackle the generalization challenge, two primary paradigms have emerged: scaling real-world data collection and synthetic data augmentation. Inspired by scaling laws in foundation models~\cite{gpt4o,yang2025qwen3,ravi2025sam}, the former relies on large-scale pretraining~\citep{o2024open,wu2024robomind,hou2025robomind,wu2025robocoin}. However, unlike internet-scale text and image data, collecting robotic demonstrations is costly and labor-intensive. As a result, data augmentation (DA)~\cite{lee2019network,hansen2021generalization,Yu-RSS-23,liu2025DCODA} has become a practical alternative. Traditional ``weak'' augmentations, such as random cropping and color jittering, alter low-level pixel statistics but lack the semantic diversity needed to bridge the gap between training and deployment environments.


Recent works~\cite{Chen-RSS-23,wang2024cyberdemo,chen2025semantically,11247097,zhao2025efficient} have advanced ``strong'' semantic augmentation by using generative models~\cite{ramesh2021zero,rombach2022high} to synthesize diverse visual contexts via inpainting, but they heavily depend on accurate segmentation~\cite{ravi2025sam} or detection~\cite{oquab2024dinov2} of task-relevant entities. As observed in RoboEngine~\cite{11247097} and verified by our experiments on \textbf{RoboAug-D}, a dataset of 7,576 trajectories across 33 tasks, state-of-the-art detectors such as GroundingDINO~\cite{liu2024grounding}, LLMDet~\cite{fu2025llmdet}, and SAM3 ~\cite{carion2026sam} frequently fail under real-world conditions. These errors propagate through the generation pipeline, where missed or hallucinated regions corrupt inpainted scenes and lead policies to learn incorrect behaviors, such as grasping empty space. Consequently, existing methods struggle to generate the high-quality augmented data required for robust real-world generalization.


To overcome these deficiencies, we propose \textbf{RoboAug}, a Region-Contrastive Data Augmentation Framework for robust generalization with minimal human supervision. RoboAug integrates three stages: task-relevant region extraction, semantic augmentation, and region-contrastive policy learning. Given annotations from only a single frame, RoboAug generates semantic masks for entire trajectories through a training-free one-shot matching pipeline. By combining GroundingDINO for proposal generation, DINOv2~\cite{oquab2024dinov2} for category matching, and SAM2~\cite{ravi2025sam} for temporal tracking, our method achieves accurate pixel-level extraction of task-relevant entities without frame-by-frame annotation or detector retraining~\cite{11247097}.

Building on these high-quality masks, RoboAug performs semantic augmentation by synthesizing diverse full-scene backgrounds~\cite{yeh2017semantic} and compositing foreground regions onto them, avoiding the instability of inpainting-based methods~\cite{wang2023imagen}. To further leverage mask semantics, we introduce a region-contrastive objective that directly applies contrastive learning to the visual encoder without architectural changes, encouraging semantic clustering while separating irrelevant features. This improves the policy’s focus on task-relevant objects under visual interference.

We validate RoboAug through over \textbf{35k} real-world trials on Tian Gong 2.0, UR-5e, and AgileX. Our evaluation systematically isolates and combines background, lighting, and distractor variations to assess robustness under diverse OOD conditions. In the most challenging setting involving all three factors, RoboAug achieves success rates of 0.67, 0.47, and 0.60, substantially outperforming the strongest baseline (0.42, 0.31, and 0.34). These results demonstrate the importance of combining accurate region extraction with contrastive learning for robust robot learning.
Our main contributions are summarized as follows:
\vspace{-5pt}
\begin{itemize}
    \item We propose RoboAug, a region-contrastive data augmentation framework that facilitates the scalable generation of diverse training data with minimal human supervision.
    
    \item We introduce a one-shot region matching strategy combined with a region-contrastive loss, significantly improving both the precision of visual extraction and the expressiveness of learned policy features.
    
    \item Through extensive real-world experiments exceeding 35k trials, RoboAug exhibits robust generalization against diverse visual perturbations, outperforming baselines by 59.5\%, 51.6\%, and 76.4\% on Tian Gong 2.0, UR-5e, and AgileX, respectively.
    
    \item We will open-source the embodied object detection dataset and our multi-task real-world manipulation dataset to facilitate further research.
\end{itemize}

\section{Related Work}

\subsection{Generalization in Visuomotor Policy Learning}

Generalizing visuomotor policies to unstructured environments remains a central challenge. Early imitation learning (IL) methods were limited by demonstration diversity~\citep{cui2022play,zhao2023learning,gervet2023act3d,chi2023diffusion,pearce2023imitating_diffusion,reuss2023goal,wu2024discrete,ze20243d,ke20243d,fu2024mobile,cao2024mamba,su2025dense}, whereas Vision-Language-Action (VLA) models improve generalization through large-scale cross-embodiment training. Models such as RT-1~\citep{Brohan-RSS-23}, RT-2~\citep{zitkovich2023rt}, RT-X~\citep{o2024open}, and Octo~\citep{team2024octo}, trained on datasets~\citep{khazatsky2024droid,bu2025agibot,jiang2025galaxea,wu2025robocoin} like BridgeData~\cite{ebert2021bridge, walke2023bridgedata}, Open X-Embodiment~\cite{o2024open}, RoboMIND~\cite{wu2024robomind,hou2025robomind}, show markedly stronger robustness. More capable architectures, from PaLM-E~\citep{driess23palme} and $\pi_{0}$~\citep{rss2025_avisionlanguagea} to recent VLA models~\citep{cheang2024gr,wang2024scaling,liu2024rdt,wen2025tinyvla,bjorck2025gr00t,li2025switchvla,fan2025diffusion,liu2025mla,yuan2026seeing,zhao2025cot,ICLR2026_f81332c0,qu2025spatialvla,wen2025diffusionvla} such as HybridVLA~\citep{liu2026hybridvla}, X-VLA~\citep{zheng2026xvla}, and XR-1~\citep{fan2025xr}, further exploit internet-scale data for stronger reasoning and instruction following.
Yet collecting robotic interaction data is far costlier than gathering internet-scale language or vision data, so existing datasets still fail to capture real-world visual diversity. To close this gap without expensive collection, we propose RoboAug, an architecture-agnostic data-level augmentation framework that integrates seamlessly into diverse visuomotor policies and VLA models to improve robustness against environmental variation.

\subsection{Data Augmentation for Robotic Manipulation}

Data augmentation offers a practical alternative to costly large-scale robotic data collection. Weak augmentations such as random cropping and noise injection improve robustness to low-level perturbations but lack the semantic diversity needed for out-of-distribution generalization, which has motivated strong generative augmentation~\cite{mandi2022cacti,Chen-RSS-23,YangS-RSS-25} using diffusion models. GenAug~\cite{Chen-RSS-23} first retargeted behaviors to unseen environments through text-guided background inpainting while keeping robot poses fixed, and later work extended this to distractor synthesis, action-aware augmentation, and cross-view or cross-embodiment adaptation~\cite{wang2024cyberdemo,teoh2024green,tian2024viewinvariant,chen2025semantically}, including ROSIE~\cite{Yu-RSS-23}, RoboAgent~\cite{bharadhwaj2024roboagent}, Mirage~\cite{chen2024mirage}, and RoVi-Aug~\cite{chen2024roviaug}. A central challenge is preserving task-relevant entities, since inaccurate masking corrupts critical geometric cues. To avoid restrictive setups such as green screens~\cite{teoh2024green}, recent methods adopt automated segmentation~\cite{11246305,11247097,liu2025DCODA}, yet those built on general-purpose models like SAM2~\cite{ravi2025sam} often fail under occlusion. RoboAug instead uses a one-shot task-relevant region extraction pipeline with a region-contrastive objective that enforces representation invariance on manipulated objects while improving robustness to background variation.

\section{Methodology}

\begin{figure*}[t]
    \centering
    \includegraphics[width=0.98\linewidth]{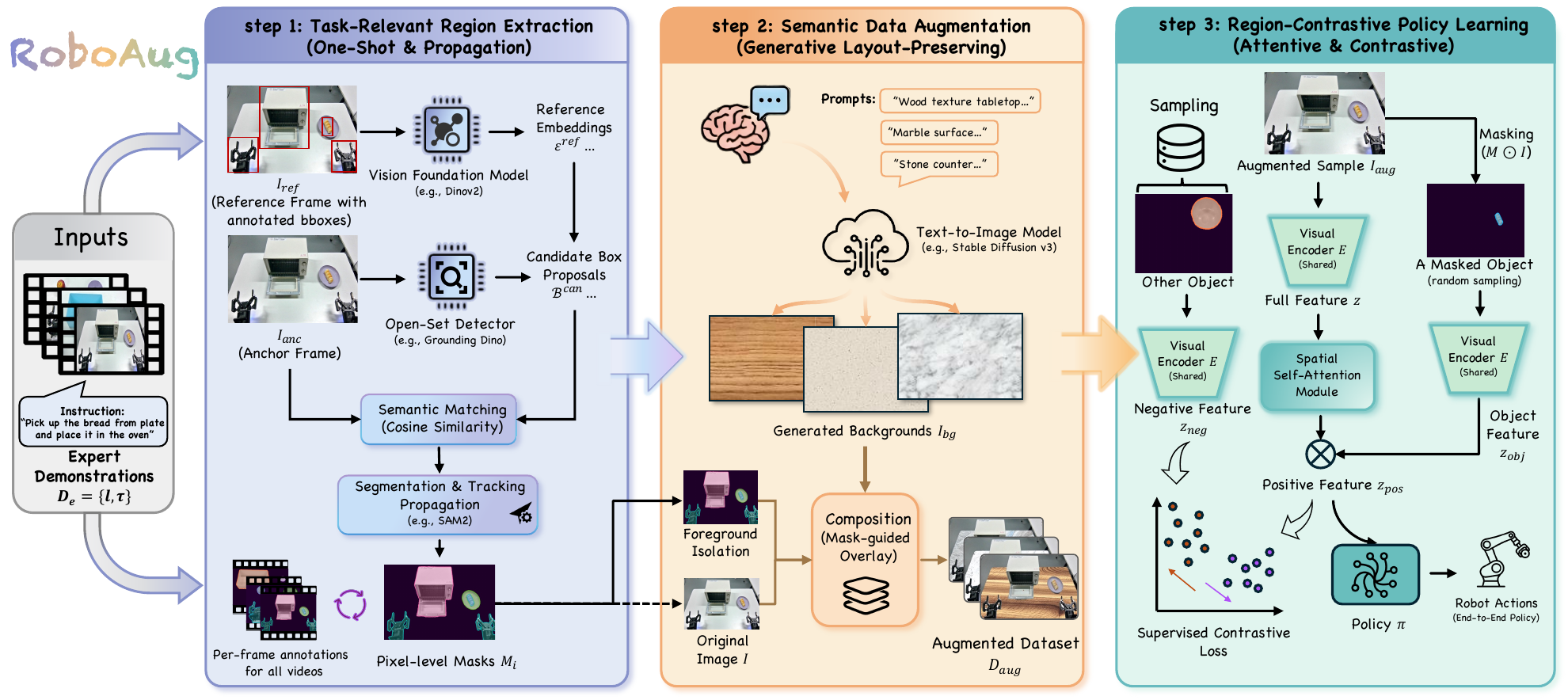}
    \caption{\textbf{Overview of RoboAug.} RoboAug contains three stages: (1) task-relevant region extraction, (2) semantic data augmentation, and (3) region-contrastive policy learning.}
    \label{fig:overview}
\end{figure*}

\subsection{Overview}

We study generalization in real-world robotic manipulation via single-task imitation learning. Given a task described by a language instruction $l$, we collect an expert dataset $\mathcal{D}_{e} = \{(l, \tau_i)\}_{i=1}^{N}$, where each trajectory $\tau = \{(o_t, m_t, a_t)\}_{t=1}^{T}$ records camera images $o_t$, proprioceptive states $m_t$, and robot actions $a_t$. We split each observation $o$ into a task-relevant region $R_{\text{task}}$, including the robot arm and all entities needed to determine the demonstrated action (e.g., manipulated objects, destination containers, and
task-relevant obstacles), and a task-irrelevant scenario region $R_{\text{scen}}$. Our goal is to learn a visuomotor policy $\hat{a}_t = \pi(o_t, m_t)$ that keeps a high success rate in novel environments with unseen scenarios $R_{\text{scen}}^{\text{new}}$ while $R_{\text{task}}$ remains invariant. To this end, we propose \textbf{RoboAug}, a region-contrastive data augmentation framework that improves semantic diversity and feature robustness at low annotation cost. As shown in Figure~\ref{fig:overview}, RoboAug extracts $R_{\text{task}}$, applies semantic augmentation to $R_{\text{scen}}$, and trains the policy with a region-contrastive objective, guiding it to ground decisions in task-critical features for robust generalization. Implementation details and informal motivation are provided in Appendix~\ref{supp:imple} and Appendix~\ref{supp:theory}.

\subsection{Task-Relevant Region Extraction}

This stage aims to obtain pixel-level masks $M_{\text{task}} \in {0, 1}^{H \times W}$ for task-relevant regions $R_{\text{task}}$ from demonstration trajectories $\tau$ without extensive annotation or detector retraining. To achieve this, we introduce a lightweight two-step pipeline requiring only one manually labeled reference image per task. We first localize task-relevant elements in each trajectory’s anchor frame through training-free one-shot region matching, then propagate the annotations across subsequent frames via segmentation and tracking to produce temporally consistent pixel-level masks.

\textbf{One-Shot Region Matching}.
We can designate an arbitrary frame from one trajectory in which the task-relevant regions are clearly visible as the \textit{reference frame} $I_{\text{ref}}$, and manually annotate bounding boxes $\mathcal{B}^{\text{ref}} = \{B^{\text{ref}}_{i}\}_{i=1}^{K}$ for the $K$ task-relevant regions (e.g., manipulated objects). These regions are then cropped and encoded into reference embeddings $\mathcal{E}^{\text{\text{ref}}} = \{ \bcode^{\text{ref}}_i \}_{i=1}^K$ using a vision foundation model (DINOv2~\cite{oquab2024dinov2} in our implementation), where $\bcode^{\text{ref}}_i \in \mathbb{R}^d$.
To transfer these labels to the remaining trajectories, we treat the first frame of each subsequent trajectory $\tau$ as the \textit{anchor frame}, denoted as $I_{\text{anc}}$. 
For each $I_{\text{anc}}$, we utilize an open-set detector (GroundingDINO~\cite{liu2024grounding} in our implementation) to generate candidate bounding box proposals $\mathcal{B}^{\text{can}} = \{B^{\text{can}}_{j}\}_{j=1}^{M}$.
For each candidate $B_{j}^{\text{can}}$, we extract its feature embedding $\bcode_{j}^{\text{can}}$ and measure its semantic alignment with the reference templates $\{\bcode_{i}^{\text{ref}}\}_{i=1}^K$ via cosine similarity. The predicted category $\hat{c}_j$ is determined by the most similar reference embedding:
\begin{equation}
    \hat{c}_j = \operatorname*{argmax}_{i \in \{1, \dots, K\}} \text{sim}(\bcode_{j}^{\text{can}}, \bcode_{i}^{\text{ref}}).
    \label{similarity}
\end{equation}
This mechanism ensures robust, training-free alignment of task-relevant regions across diverse demonstrations while filtering out irrelevant background clutter.

\textbf{Semantic Mask Propagation}.
After identifying task-relevant bounding boxes and categories in each anchor frame $I_{\text{anc}}$, we extend this semantic information to the full trajectories $\{ \tau_i \}_{i=1}^{N}$ via SAM-2~\cite{ravi2025sam}, which transforms sparse bounding box priors $\mathcal{B}^{\text{can}}$ into dense pixel-level masks $M_{\text{task}}$ and propagates them with spatiotemporal consistency. This yields per-frame semantic masks while requiring manual bounding box annotations for only a single reference frame per task. These masks serve a dual purpose: (1) precisely delineating task-relevant regions for downstream semantic data augmentation via generative models, and (2) providing fine-grained supervision that enhances feature representation in region-contrastive policy learning.



\subsection{Semantic Data Augmentation }


Given the pixel-level masks $M_{\text{task}}$ for task-relevant regions, RoboAug
performs semantic augmentation that diversifies the training data through
environmental variation while preserving task-critical structure. We prompt a large
language model (ChatGPT~\cite{chatgpt}) for descriptive captions that guide image
synthesis, drawing from a library of 500 background templates spanning wood
($58\%$), stone ($35\%$), and composite ($7\%$) materials common to tabletop scenes.



Unlike prior methods based on inpainting, we synthesize a complete background independently and composite the original task-relevant foreground onto it. Let $I$ be the
original image and $M_{\text{task}}$ the mask of task-relevant regions, such as the
robot arm and manipulated objects. We synthesize a full-frame background
$I_{\text{bg}}$ with Stable Diffusion v3~\cite{esser2024scaling}, then composite the
preserved foreground onto it:
\begin{equation}
    I_{\text{aug}} = M_{\text{task}} \odot I + (1 - M_{\text{task}}) \odot I_{\text{bg}},
\end{equation}
where $\odot$ denotes element-wise multiplication. By construction, pixels within $M_{\text{task}}$ are preserved
exactly, preventing the background generator from altering
the retained robot and object regions. Repeating this with a randomly
sampled prompt per image yields the augmented dataset
$\mathcal{D}_{\text{fnl}} = \mathcal{D}_{e} \cup \mathcal{D}_{\text{aug}}$, which
contains thousands of distinct backgrounds that simulate varied environments while
keeping task-relevant regions intact.


\subsection{Region-Contrastive Policy Learning}



Prior data augmentation techniques enhance diversity but often neglect task-relevant regional semantics during policy training. To address this, we introduce a Region-Contrastive Learning (RCL) objective that leverages these regions to refine the policy's visual representations.
During training, for each image $I$ sampled from the final dataset $\mathcal{D}_{\mathrm{fnl}}$, we construct
masked images corresponding to its task-relevant regions. Given the binary mask $M_{\mathrm{task},c}$ associated
with region identity $c$, we obtain
$I_{\mathrm{obj}} = M_{\mathrm{task},c} \odot I$. Here, $c$ identifies an annotated task-relevant region rather than the task itself.
The full and masked images are processed by a shared visual encoder $E(\cdot)$ to obtain $z = E(I)$ and $z_{\mathrm{obj}} = E(I_{\mathrm{obj}})$, respectively. Both are spatial feature maps in
$\mathbb{R}^{B \times C \times H' \times W'}$,
where $B$ is the image batch size, $C$ is the number of feature channels, and $H'$ and $W'$ are the spatial dimensions.

However, the masked images $I_{\mathrm{obj}}$ are often
dominated by zero-valued regions, which can dilute
task-relevant information in their feature representations.
To mitigate this, we use the full-image features $z$ to
compute spatial weights for $z_{\mathrm{obj}}$.
Specifically, a spatial attention module aggregates $z$
along the channel dimension using maximum and average
pooling, concatenates the resulting maps, and applies a
learnable convolution followed by a sigmoid:
\begin{equation}
    a_{\mathrm{att}}
    =
    \sigma\left(
        \mathcal{A}
        \left(
            [\operatorname{Max}_{c}(z);
             \operatorname{Avg}_{c}(z)]
        \right)
    \right),
    \qquad
    z_{\mathrm{att}}
    =
    a_{\mathrm{att}} \odot z_{\mathrm{obj}}.
    \label{eq:region_attention}
\end{equation}
Here, $\mathcal{A}$ denotes the learnable convolution,
$[\cdot;\cdot]$ denotes channel-wise concatenation, and
the subscript $c$ on the pooling operators denotes the
channel dimension.
The attention map
$a_{\mathrm{att}} \in
\mathbb{R}^{B \times 1 \times H' \times W'}$
is broadcast across channels when multiplied with
$z_{\mathrm{obj}}$.
The same attention map also gates the full-image features
passed to the policy, yielding $a_{\mathrm{att}} \odot z$.

To align representations of the same region while
separating different regions, we optimize the visual
encoder using a supervised contrastive objective inspired by~\cite {khosla2020supervised}.
For each region sample $i$, we flatten its attended
feature map and apply $\ell_2$ normalization to obtain
$u_i = \operatorname{Normalize}
(\operatorname{vec}(z_{\mathrm{att},i}))
\in \mathbb{R}^{d}$, where $d = C H' W'$.
Let $\mathcal{B}_{\mathrm{obj}}$ denote the index set of
region samples in a mini-batch, with $c_i$ denoting the
region identity of sample $i$.
For each anchor $i$, we define
$S(i) = \mathcal{B}_{\mathrm{obj}} \setminus \{i\}$ and
$P(i) = \{p \in S(i) : c_p = c_i\}$.
Other samples of the same region form positive pairs,
whereas samples with different region identities form
negative pairs.
Thus, a single-task setting does not imply a single
contrastive label: multiple annotated regions within the
same task provide distinct identities, allowing both
positive and negative pairs to be formed without
requiring samples from different tasks.

The region-contrastive loss is defined as
\begin{equation}
    \mathcal{L}_{\mathrm{RC}}
    =
    -\frac{1}{|\mathcal{B}_{\mathrm{obj}}|}
    \sum_{\substack{
        i \in \mathcal{B}_{\mathrm{obj}}\\
        |P(i)| > 0
    }}
    \frac{1}{|P(i)|}
    \sum_{p \in P(i)}
    \log
    \frac{
        \exp(u_i^\top u_p / \tau)
    }{
        \sum_{j \in S(i)}
        \exp(u_i^\top u_j / \tau)
    },
    \label{eq:region_contrastive}
\end{equation}
where $\tau$ denotes the temperature.
The anchor itself is excluded from both the positive
set and the denominator, while the denominator includes
all other samples, comprising both positives and negatives.
Anchors without positive samples contribute zero to
the loss.
We jointly optimize the policy and visual encoder with
$\mathcal{L}_{\mathrm{total}}
= \mathcal{L}_{\mathrm{action}}
+ \lambda \mathcal{L}_{\mathrm{RC}}$,
where $\mathcal{L}_{\mathrm{action}}$ is the
policy-specific imitation-learning objective and
$\lambda$ balances the two terms.
Masked region inputs are required only during training;
at inference, the policy uses the full-image branch
with its spatial attention module. An ablation of the attention design is provided in
Appendix~\ref{sec:additional_ablations}.

\subsection{RoboAug-D Dataset for Object Detection}


State-of-the-art vision foundation models, such as GroundingDINO~\cite{liu2024grounding}, LLMDet~\cite{fu2025llmdet}, and SAM3~\cite{carion2026sam} often exhibit performance degradation when applied to robotic manipulation scenes. 
\begin{wraptable}{r}{0.48\textwidth}   
    \centering
    \vspace{-5pt}
    \caption{RoboAug-D Dataset Statistics.}
    \label{tab:roboaug-d}
    \resizebox{\linewidth}{!}{
    \begin{tabular}{c|ccccc}
        \toprule
        Robot & Task & Traj. & Frame & Obj. & BBox. \\ 
        \midrule
        Single-Arm Franka & 8 & 2442 & 20511 & 19 & 87426 \\
        Single-Arm UR & 17 & 4217 & 40136 & 34 & 197882  \\
        Dual-Arm UR & 3 & 669 & 11077 & 11 & 71464 \\
        Dual-Arm Agilex & 5 & 248 & 2025 & 9 & 10063  \\
        \midrule
        Total & 33 & 7576 & 73749 & 46 & 366835 \\
        \bottomrule
    \end{tabular}%
    }
\end{wraptable}
To investigate and benchmark model robustness in these domains, we introduce \textbf{RoboAug-D}, a large-scale object detection dataset manually annotated from the perspective of robotic manipulators. 
As shown in Table~\ref{tab:roboaug-d}, the dataset encompasses 33 distinct tasks, comprising a total of 73,749 keyframes and 366,835 bounding boxes across 46 object categories.

\section{Experiments}

We evaluate RoboAug along a progression from basic visual perception to full physical manipulation.
Section~\ref{exp:obj_det} first probes the limits of state-of-the-art vision foundation models.
Moving to the physical world, Sections~\ref{exp:setup} and ~\ref{exp:compos} report manipulation experiments that measure both broad compositional generalization and single-factor robustness.
Section~\ref{sec:rcl_effect} then ablates the performance of region-contrastive learning. Additional experiments, including simulation, scaling laws regarding augmentation magnitude, and multi-view augmentation, are deferred to Appendices~\ref{supp:dual}--\ref{sup:multi-view}.


\subsection{Object Detection on RoboAug-D Dataset}
\label{exp:obj_det}


\textbf{Experimental Setup.} 
We evaluated the zero-shot object detection capabilities of Vision Foundation Models (VFMs) on the full test set of the RoboAug-D dataset, with a focus on challenges inherent to robotic manipulation scenarios.
Our detection results are obtained through training-free reference matching: frozen GroundingDINO generates candidate proposals, and frozen DINOv2 features are used to match them to separately annotated reference images. The policy's visual encoder is not involved in this
evaluation, and no model is fine-tuned on RoboAug-D. The competing VFMs are evaluated in their frozen zero-shot settings. Thus, this comparison evaluates reference-guided localization against zero-shot detection, with the additional reference supervision explicitly provided to our method. And reference images are excluded from the evaluated test set.
For each of the 46 object categories, we queried the models using five diverse text prompts (e.g., varying in phrasing, synonyms, and functional descriptions).
Model performance was compared using mean average precision (mAP@0.5), and our proposed approach was benchmarked against GroundingDINO, LLMDet and SAM3.

\begin{figure}[!t]
    \centering
    \includegraphics[
        width=0.9\linewidth,
        trim = 5mm 6mm 1mm 5mm,  
        clip
    ]{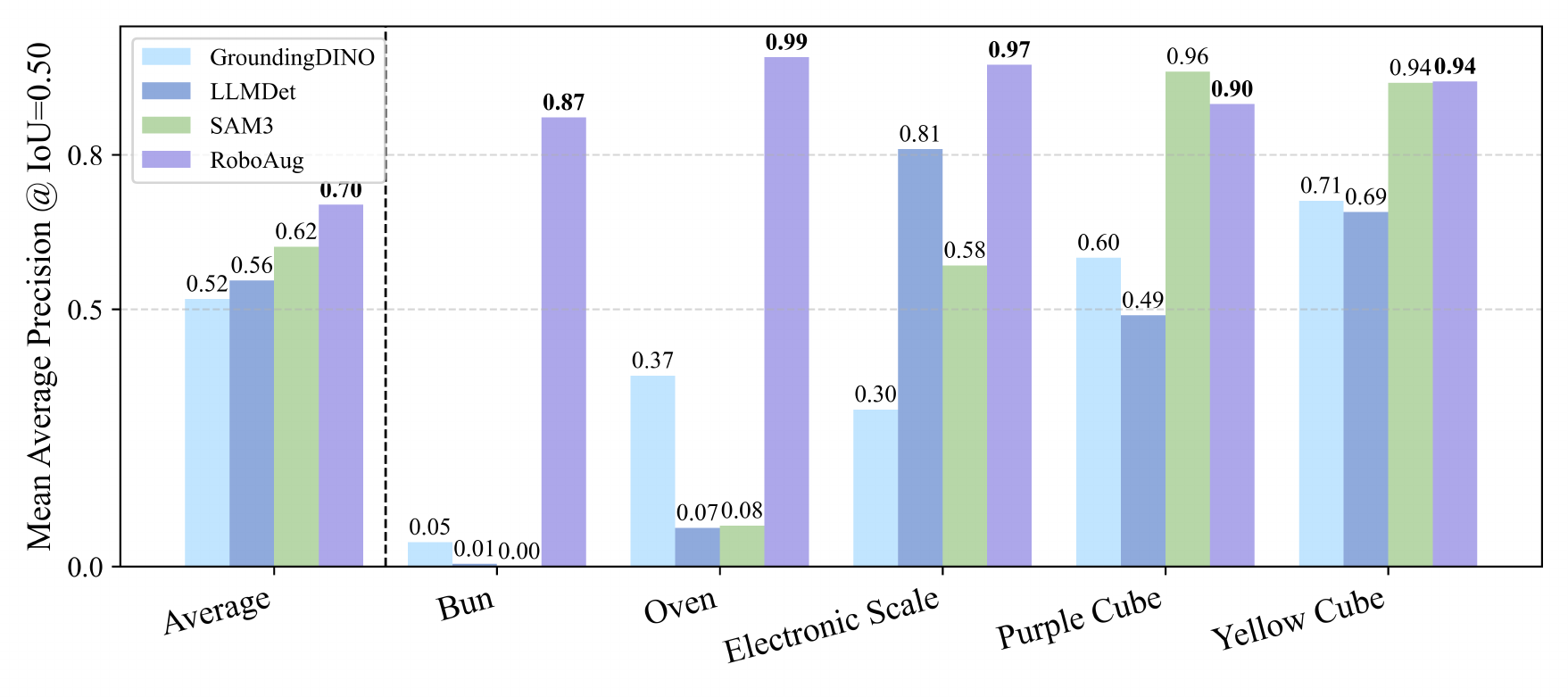}
    \caption{Comparison of mAP@0.5 across RoboAug-D Dataset. We present the results of 5 representative objects.}
    \label{fig:visual_test}
\end{figure}

\textbf{Experimental Results.} 
Figure~\ref{fig:visual_test} provides the quantitative results, detailing both the overall mAP@0.5 on representative object categories. 
Our approach demonstrates significant improvements, outperforming the state-of-the-art baselines GroundingDINO, LLMDet and SAM3 by 34.6\%, 25.0\% and 12.9\%, respectively. 
For instance, in the ``Bun'' category, baselines struggle to exceed 0.10 mAP, whereas our method achieves scores of 0.87. 
These results highlight the limitations of current VFMs in handling robotic viewpoints and validate the effectiveness of our one-shot region matching strategy in enhancing detection accuracy for downstream data augmentation.

\subsection{Real-World Generalizable Robotic Manipulation}
\label{exp:setup}


\begin{figure*}[!t]
    \centering
    \includegraphics[width=0.96\linewidth]{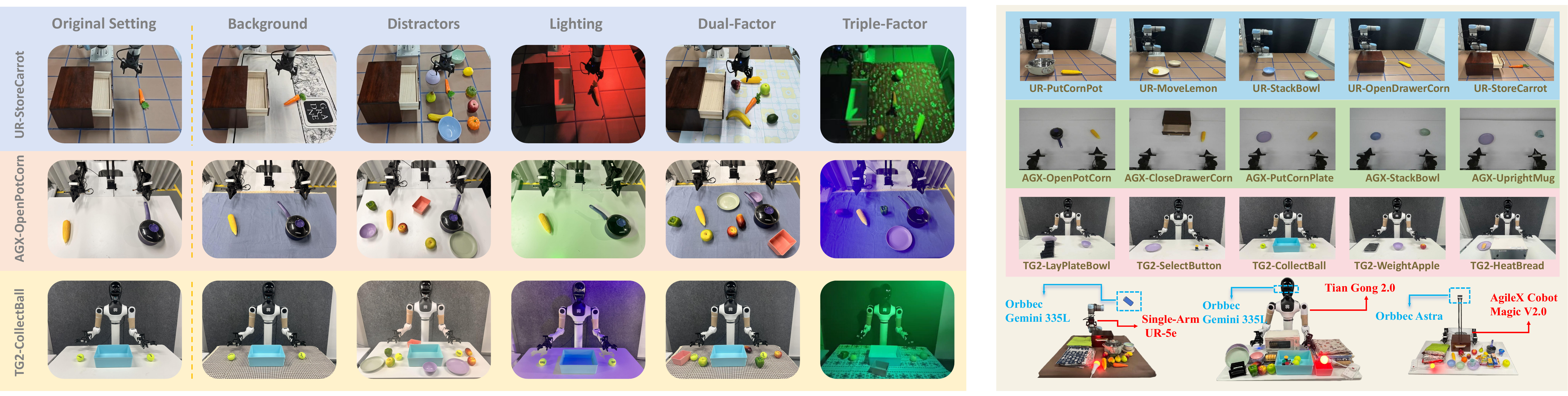}
    \caption{Left: Overview of the generalization evaluation settings; Right: Experimental setup.}
    \label{fig:test_scenario}
\end{figure*}

\textbf{Hardware Setup.}
We validate RoboAug across diverse robots illustrated in Figure~\ref{fig:test_scenario}: 
(1) the single-arm UR-5e, 
(2) the Tian Gong 2.0 humanoid robot, 
and (3) the AgileX Cobot Magic V2.0 robot. 
And we collect the dataset via human teleoperation HACTS~\cite{xu2025hacts}, recording visual observations, robot states, and actions at every frame.

\textbf{Task Design.}
As shown in Figure~\ref{fig:test_scenario}, we designed five tasks per embodiment to cover a range of complexities, extending from single-arm pick-and-place to precise dual-arm collaboration. 
These tasks require diverse skills, including pushing, rotating, and grasping. 
The UR-5e performs household interactions such as PutCornPot and OpenDrawerCorn. 
The AgileX and Tian Gong 2.0 robots execute complex bimanual tasks, including UprightMug and LayPlateBowl. 
For each task, we collected a dataset comprising 50 expert trajectories.

\textbf{Generalization Evaluation and Metrics.}
We devised two protocols to assess robustness:
\textit{Compositional Generalization}, which evaluates adaptability across combined environmental variables, and \textit{Single-Factor Generalization}, which probes stability against intense variations in specific factors. 
Variables include background textures, lighting conditions, and task-irrelevant distractors. 
We report the success rate averaged over 20 real-world rollouts per configuration.

\subsection{Compositional Generalization Evaluation}
\label{exp:compos}
\textbf{Evaluation Setup.}
We evaluate our policy under a challenging \textit{triple-factor} protocol incorporating 3 unseen backgrounds, 3 task-irrelevant distractors, and 4 distinct lighting conditions. 
We compared RoboAug against a non-augmented policy (ACT~\cite{zhao2023learning}), a texture-replacement method (RoboEngine-T~\cite{11247097}), and two generative baselines (RoboEngine-G~\cite{11247097} and GenAug~\cite{Chen-RSS-23}). 
All augmentation methods employ a $5\times$ data expansion ratio, supplementing 50 real-world expert trajectories with 250 generated trajectories. 
Additional results of the \textit{Dual-Factor} setting are detailed in Appendix~\ref{supp:dual}.

\textbf{Results under Triple-Factor Variation.}
Table~\ref{tab:three_main_all} shows the performance on all the robots. 
RoboAug consistently outperforms all baselines. 
Notably, in the AGX-UprightMug task, which requires rotating a mug and coordinating placement, RoboAug achieves a success rate of 0.58, significantly surpassing the strongest baseline GenAug (0.34). 
Results for UR-5e and Tian Gong 2.0 show similar gains, demonstrating the embodiment-agnostic generalization of RoboAug.

\begin{wraptable}{r}{0.50\textwidth}   
    \centering
    \vspace{-1pt}
    \caption{Quantitative Results on $\pi_{0}$.}
    \label{tab:rcl_pi0}
    \resizebox{\linewidth}{!}{
        \begin{tabular}{l|cc|c}
            \toprule
            \textbf{Method} & UR-PutCornPot & UR-MoveLemon & \textbf{Average} \\
            \cmidrule(lr){1-4} 
            $\pi_0$  & 0.30 & 0.35 & - \\
            $\pi_0$ w/ RoboAug & 0.65 & 0.70 & - \\
            \midrule
            \textbf{Method} & TG2-CollectBall & TG2-HeatBread & \textbf{Average}\\
            \cmidrule(lr){1-4} 
            $\pi_0$ & 0.25 & 0.30 & 0.28 \\
            $\pi_0$ w/ RoboAug  & 0.60 & 0.65 & 0.65 \\
            \bottomrule
        \end{tabular}%
    \vspace{-5pt}
    }
\end{wraptable}
To verify the universality of our proposed method, we also integrate RoboAug into $\pi_{0}$, a model pretrained on a large-scale dataset. As illustrated in Table~\ref{tab:rcl_pi0}, the experimental results demonstrate that our framework consistently improves performance for any policies.
These findings validate the effectiveness of RoboAug and suggest that it is compatible with diverse policy formulations.



\textbf{Single-Factor Generalization Evaluation.}
To rigorously assess generalization boundaries, we isolate three environmental factors: 
(1) \textit{Background Diversity}, where we introduce 170 unseen textures across three complexity levels (Geometric, Structured, Scattered); 
(2) \textit{Distractor Density}, where we increase workspace clutter to 10 objects; 
and (3) \textit{Lighting}, which spans 20 distinct conditions including dynamic shifts. 
We provide results in~\Cref{supp:distractor,supp:light} of the Appendix.


\subsection{Effectiveness of Region-Contrastive Learning}
\label{sec:rcl_effect}


\begin{table*}[!t]
  \centering
  \caption{Comparative results under triple-factor variations: 3 unseen backgrounds, 4 lighting conditions, and 3 distractors.
  }
  \resizebox{0.98\textwidth}{!}{
  \begin{tabular}{l|ccccc|c}
    \toprule
    \textbf{Augmentation Method} & UR-PutCornPot & UR-MoveLemon & UR-StackBowl & UR-StoreCarrot & UR-OpenDrawerCorn & \textbf{Average}\\
    \cmidrule(lr){1-7}
    ACT w/o Aug~\cite{zhao2023learning}  & 0.06 & 0.06 & 0.08 & 0.10 & 0.15 & 0.09 \\
    RoboEngine-T~\cite{11247097}  & 0.12 & 0.16 & 0.12 & 0.20 & 0.32 & 0.18 \\
    RoboEngine-G~\cite{11247097}  & 0.16 & 0.24 & 0.12 & 0.32 & 0.40 & 0.25 \\
    GenAug~\cite{Chen-RSS-23}  & 0.22 & 0.28 & 0.16 & 0.40 & 0.48 & 0.31 \\
    \textbf{RoboAug} & \textbf{0.38} & \textbf{0.46} & \textbf{0.28} & \textbf{0.56} & \textbf{0.68} & \textbf{0.47} \\
    \midrule
     & AGX-PutCornPlate & AGX-UprightMug & AGX-StackBowl & AGX-OpenPotCorn & AGX-CloseDrawerCorn &  \\
    \cmidrule(lr){1-7}
    ACT w/o Aug~\cite{zhao2023learning}  & 0.12 & 0.14 & 0.14 & 0.16 & 0.24 & 0.16 \\
    RoboEngine-T~\cite{11247097}  & 0.24 & 0.22 & 0.28 & 0.30 & 0.28 & 0.26 \\
    RoboEngine-G~\cite{11247097}  & 0.30 & 0.32 & 0.32 & 0.28 & 0.38 & 0.32 \\
    GenAug~\cite{Chen-RSS-23}  & 0.32 & 0.34 & 0.30 & 0.32 & 0.40 & 0.34 \\
    \textbf{RoboAug} & \textbf{0.51} & \textbf{0.58} & \textbf{0.62} & \textbf{0.55} & \textbf{0.73} & \textbf{0.60} \\
    \midrule
    & TG2-WeightApple & TG2-CollectBall & TG2-HeatBread & TG2-SelectButton & TG2-LayPlateBowl &  \\
    \cmidrule(lr){1-7}
    ACT w/o Aug~\cite{zhao2023learning}   & 0.20 & 0.16 & 0.14 & 0.22 & 0.24 & 0.19 \\
    RoboEngine-T~\cite{11247097}  & 0.28 & 0.25 & 0.32 & 0.34 & 0.30 & 0.30 \\
    RoboEngine-G~\cite{11247097}  & 0.38 & 0.36 & 0.40 & 0.35 & 0.42 & 0.38 \\
    GenAug~\cite{Chen-RSS-23}  & 0.48 & 0.32 & 0.42 & 0.40 & 0.50 & 0.42 \\
    \textbf{RoboAug} & \textbf{0.80} & \textbf{0.52} & \textbf{0.65} & \textbf{0.68} & \textbf{0.70} & \textbf{0.67} \\
    \bottomrule
  \end{tabular}
  }
  \label{tab:three_main_all}
\end{table*}


\begin{table*}[!t]
  \centering
  \caption{Ablation study results on region-contrastive loss (RCL).}
  \resizebox{0.75\textwidth}{!}{
  \begin{tabular}{l|c|ccc|c}
    \toprule
    \textbf{Method} & \textbf{RCL} & UR-PutCornPot & AGX-PutCornPlate & TG2-WeightApple & \textbf{Average} \\
    \midrule
    ACT w/o Aug~\cite{zhao2023learning} & $\times$ & 0.06 & 0.12 & 0.20 & 0.13 \\
    ACT w/o Aug~\cite{zhao2023learning} & $\checkmark$ & 0.12 & 0.20 & 0.32 & 0.21 \\
    \cmidrule(lr){1-6}
    RoboEngine-T~\cite{11247097} & $\times$ & 0.12 & 0.24 & 0.28 & 0.21 \\
    RoboEngine-T~\cite{11247097} & $\checkmark$ & 0.14 & 0.28 & 0.30 & 0.24 \\
    \cmidrule(lr){1-6}
    RoboEngine-G~\cite{11247097} & $\times$ & 0.16 & 0.30 & 0.38 & 0.28 \\
    RoboEngine-G~\cite{11247097} & $\checkmark$ & 0.16 & 0.36 & 0.44 & 0.32 \\
    \cmidrule(lr){1-6}
    GenAug~\cite{Chen-RSS-23} & $\times$ & 0.22 & 0.32 & 0.48 & 0.34 \\
    GenAug~\cite{Chen-RSS-23} & $\checkmark$ & 0.28 & 0.35 & 0.53 & 0.39 \\
    \cmidrule(lr){1-6}
    \textbf{RoboAug} & $\times$ & 0.28 & 0.43 & 0.68 & 0.46 \\
    \textbf{RoboAug} & $\checkmark$ & \textbf{0.38} & \textbf{0.51} & \textbf{0.80} & \textbf{0.56} \\
    \bottomrule
  \end{tabular}
  }
  \label{tab:rcl_ablation}
\end{table*}

\textbf{Quantitative Results.}
Table~\ref{tab:rcl_ablation} presents the impact of the region-contrastive loss on policy performance. 
We observe a consistent improvement across all baselines when integrating RCL. 
Even for the standard baseline ACT w/o Aug, RCL doubles the success rate in the UR-PutCornPot task. 
The addition of RCL boosts RoboAug's performance on the challenging TG2-WeightApple task from 0.68 to 0.80. 
This trend indicates that RCL effectively enhances feature robustness and generalization capability, regardless of the underlying data augmentation strategy.


\begin{wrapfigure}{r}{0.48\textwidth}  
    \centering
    \vspace{-12pt}
    \includegraphics[width=\linewidth]{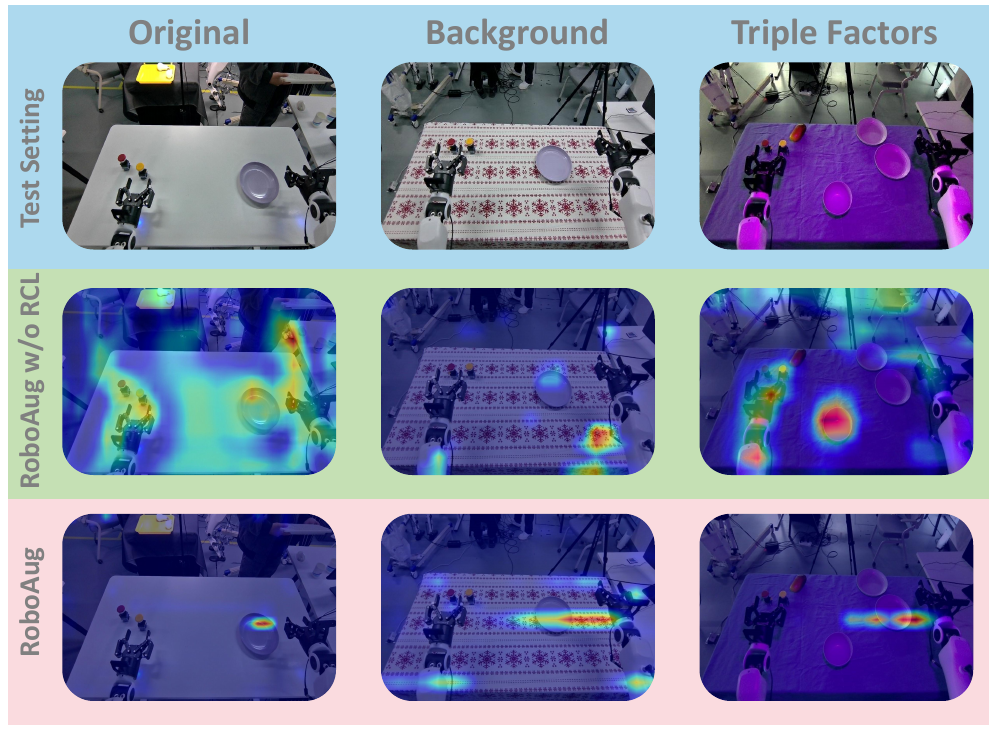} 
    \caption{Feature heatmap comparison of RoboAug with and without RCL.}
    \label{fig:heatmap}
     \vspace{-10pt}
\end{wrapfigure}

\textbf{Visualization Analysis.}
To understand what the policy learns, we visualize Grad-CAM~\cite{selvaraju2017grad}
heatmaps in Figure~\ref{fig:heatmap} under three settings of increasing
difficulty, namely the original environment, an unseen background, and a complex
scene with three perturbation factors. Without region-contrastive loss
($\mathcal{L}_{\mathrm{RCL}}$), the baseline produces diffuse attention that is distracted by background textures and task-irrelevant objects, especially
under strong lighting changes. 
With $\mathcal{L}_{\mathrm{RCL}}$, RoboAug instead
keeps a tight focus on the grasping region and suppresses such noise, showing
that the $\mathcal{L}_{\mathrm{RCL}}$ drives the policy to encode task-relevant
semantics that stay stable across visual perturbations.

\paragraph{Additional ablations.}
Additional experiments on UR-StoreCarrot and TG2-WeighApple
under triple-factor shifts examine the contributions of
background generation, RCL, and mask quality.
Combining background generation and RCL achieves a success
rate of 0.68, compared with 0.55 and 0.33 using either
component alone.
Our masks achieve the same success rate as SAM2-only masks
with fewer annotated frames per task, while oracle masks
further improve performance to 0.75.
We additionally evaluate reference-frame sensitivity and
alternative attention designs.
Detailed settings and results are provided in
Appendix~\ref{sec:additional_ablations}.

\section{Conclusion and Limitation}
In conclusion, we presented RoboAug, a data augmentation framework that enhances robotic generalization using only a single reference-frame annotation per task. By combining generative semantic augmentation with a plug-and-play region-contrastive loss, RoboAug focuses learning on task-relevant regions and achieves consistent state-of-the-art performance across over 35k real-world trials on UR-5e, AgileX, and Tian Gong 2.0 robots. 

A limitation of this work is that training and evaluation
were conducted on individual tasks separately, leaving
multi-task effectiveness unexplored.
Additionally, the scene synthesis module currently relies
on 2D generative models, which primarily capture static background variations and do not explicitly model temporal consistency or physical interactions. While we evaluate robustness to lighting changes and distractors, explicitly augmenting these factors during
training also remains unexplored.
Finally, our evaluation focuses on task success under visual distribution shifts and does not establish deployment safety, particularly in visually novel environments that introduce physical hazards. In future work, we will extend the framework to multi-task scenarios and explore integrating emerging spatial world
models to enable temporally consistent scene augmentation. We will also investigate uncertainty detection and fallback mechanisms, together with safety validation under unseen conditions.


\acknowledgments{This work was supported in part by the New Generation
Artificial Intelligence National Science and Technology Major Project (Grant No.~2025ZD0122604).

Developing robust robot learning models is a challenging
endeavor that requires extensive collaboration across
algorithm design, data collection, and real-world evaluation.
We would like to extend our deepest gratitude to the
following individuals for their invaluable help in this
work: Haonan Liu, Jianwei Guo,
Jianwei Sun, Panpan Chen, Pei Ren, Shuguang Qiao, Shiwei Jiao, Bin Zhang, Yang Gao. We also sincerely appreciate the dedication
and effort of the many contributors who assisted with
data collection, quality assurance, and robot testing.
Their collective efforts and expertise have been
instrumental in making this research possible.}


\bibliography{ref}  

\clearpage

\section{Appendix}

\subsection{Implementation Details}
\label{supp:imple}
In this section, we provide a comprehensive description of the RoboAug framework, detailing the pipeline from task-relevant region extraction to region-contrastive policy learning.

\textbf{Object Detection Details.} 
To obtain accurate bounding boxes for task-relevant elements $\mathcal{B}^{\text{ref}}$, we manually annotated the initial dataset using the X-AnyLabeling tool~\cite{X-AnyLabeling}. To be specific, annotation is performed once per task during dataset preparation by the task developer or data
collector (each reference contains an average of 4.73 bounding boxes, and based on our practical annotation experience, annotation
takes approximately one minute per reference.
This is an informal estimate rather than a measurement
from a controlled user study).
These regions were then cropped and resized to $224 \times 224$ pixels to align with the input specifications of DINOv2 (86M parameters)~\cite{oquab2024dinov2} and encoded into a set of reference embeddings $\mathcal{E}^{\text{ref}}$. 
During inference, we employ the open-set detector GroundingDINO~\cite{liu2024grounding} to generate candidate bounding boxes $B_{j}^{\text{can}}$ with both box and text thresholds set to 0.15. 
For each candidate box, we extract its DINOv2 feature and assign its region identity by matching it to the
annotated reference embeddings using Eq.(~\ref{similarity}). GroundingDINO is used for proposal generation, rather
than final identity assignment.

\textbf{Semantic Data Augmentation Details.} 
We use Stable Diffusion 3 Medium~\cite{esser2024scaling} for text-to-image generation. 
The inference steps (\texttt{num\_inference\_steps}) and guidance scale (\texttt{guidance\_scale}) are set to 30 and 10.0, respectively. 
Image augmentation is performed with a batch size of 12, and generating a trajectory of 200 frames takes approximately 0.2 GPU-hours on an NVIDIA A100 GPU.

\textbf{Region-Contrastive Policy Learning Details.} 
We apply the proposed Region-Contrastive Loss to two policy architectures: ACT~\cite{zhao2023learning} and $\pi_0$~\cite{rss2025_avisionlanguagea}. 
While ACT relies solely on third-view RGB images and robot states as input, $\pi_0$ additionally incorporates language instructions. 
For ACT, the full-image and masked-region branches share
a ResNet50 visual encoder, which is jointly trained with
the policy. We optimize
$\mathcal{L}_{\mathrm{total}}
= \mathcal{L}_{\mathrm{action}}
+ \lambda \mathcal{L}_{\mathrm{RC}}$
with $\lambda = 1$.
For $\pi_0$, we initialize from the official
$\pi_0$-base checkpoint and fully fine-tune the model,
combining its flow-matching action objective with RCL.
The additional region branch is used during training;
the deployed policy does not require online region extraction.
Further training hyperparameter details are summarized in Table~\ref{tab:hyperparams}.

\begin{table}[ht]
\caption{Implementation Details.}
\label{tab:hyperparams}
\centering
\resizebox{\columnwidth}{!}{
    \begin{tabular}{c|ll|c|ll}
    \toprule
     & Hyperparameter & Value &  & Hyperparameter & Value \\
    \midrule
    \multirow{6}{*}{\shortstack{ACT}} & Batch Size & 24 & \multirow{6}*{\shortstack{$\pi_0$}} & Batch Size & 256 \\
    & Learning Rate & 1e-4 & & Learning Rate & 5e-5 \\ 
    & Optimizer & AdamW & & Optimizer & AdamW  \\ 
    & Vision Encoder  & ResNet50 & & Vision Encoder & SigLip \\  
    & Training Objective & $\ell_1$ + KL + RCL & & Training Objective & Flow Matching + RCL\\ 
    & Training Step & 50K & & Training Step & 30K \\ 
    & temperature & 0.07 & & temperature & 0.07 \\ 
    \bottomrule
    \end{tabular}
}

\end{table}

\textbf{Real-world Task Setup.}
The real-world evaluation is conducted on three robot embodiments: UR-5e (UR), AgileX (AGX), and Tian Gong 2.0 (TG2). The evaluated tasks are detailed below.
\begin{itemize}
    \item \texttt{UR-PutCornPot}: Transporting a piece of corn into a cooking pot. The corn is randomly placed within a rectangular region of $20~\text{cm} \times 60~\text{cm}$.
    \item \texttt{UR-MoveLemon}: Relocating a lemon from a plate to a bowl. The bowl is placed stochastically within the region of $20~\text{cm} \times 20~\text{cm}$ grid.
    \item \texttt{UR-StackBowl}: Stacking bowls in a controlled manner. The position of one bowl is fixed, whereas the second bowl is uniformly sampled along a straight line of length $60~\text{cm}$ to introduce spatial variation.
    \item \texttt{UR-OpenDrawerCorn}: Opening a drawer and placing corn inside. A random orientation is assigned to the corn, with the rotation angle sampled uniformly from the interval $[-\pi/4, \pi/4]$ radians.
    \item \texttt{UR-StoreCarrot}: Placing a carrot into a drawer and closing it. The carrot is initialized with a random rotation angle sampled uniformly from $[-\pi/4, \pi/4]$ radians.
    \item \texttt{AGX-OpenPotCorn}: Opening a pot lid and placing corn inside. The pot is placed at a fixed location, while a random orientation is assigned to the corn, with the rotation angle sampled uniformly from the interval $[-\pi/4, \pi/4]$ radians.
    \item \texttt{AGX-CloseDrawerCorn}: Picking up a corn and securely closing a drawer. The initial orientation of the corn is drawn from a uniform distribution over the interval $[-\pi/4, \pi/4]$ radians.
    \item \texttt{AGX-PutCornPlate}: Placing corn pieces on a plate. The plate is placed at a fixed location, while the corn is initialized with a random rotation angle sampled uniformly from $[-\pi/4, \pi/4]$ radians.
    \item \texttt{AGX-StackBowl}: Stacking bowls into a stable configuration. The position of the blue bowl is fixed, while the green bowl is randomly sampled from a $15~\text{cm} \times 15~\text{cm}$ grid region.
    \item \texttt{AGX-UprightMug}: Restoring a tilted mug to an upright position and placing it on the plate. And the plate remains stationary, whereas the mug is sampled from a uniform distribution over a $20~\text{cm} \times 20~\text{cm}$ region.
    \item \texttt{TG2-LayPlateBowl}: Taking a plate from the rack and laying a bowl on the plate. While the plate is fixed at the same position on the rack, the bowl is randomly sampled from a $15~\text{cm} \times 15~\text{cm}$ grid region.
    \item \texttt{TG2-SelectButton}: Selecting a yellow button and placing it on the plate. The position of the plate is fixed, while the yellow and red button are randomly sampled from a $15~\text{cm} \times 15~\text{cm}$ grid region.
    \item \texttt{TG2-CollectBall}: Collecting tennis balls into the box. Two tennis balls are positioned on opposite sides of a box, with each ball randomly located within a designated $10~\text{cm} \times 10~\text{cm}$ area.
    \item \texttt{TG2-WeighApple}: Placing the apple in the bowl and weighing them together. The electronic scale and bowl are fixed, whereas the apple is randomly placed within a $10~\text{cm} \times 10~\text{cm}$ area.
    \item \texttt{TG2-HeatBread}: Putting the bread into the oven and closing the door. The bread is initialized at a random position on the plate
\end{itemize}

\textbf{Dataset Collection}.
We collected RoboAug-D dataset using the HACTS teleoperation system~\cite{xu2025hacts} on five robot embodiments: single-arm Franka, single-arm UR-5e, dual-arm UR-5e, and AgileX.
To accommodate task-specific temporal structures, we tailored the keyframe extraction strategy to each task:

\begin{itemize}
    \item For basic, short-horizon tasks (e.g., single-arm UR-5e), we annotated semantic events (initial, gripper-close, and gripper-open frames).
    \item For complex, long-horizon tasks (e.g., dual-arm UR-5e and AgileX), we employed uniform sampling at 50-frame intervals.
\end{itemize}
All keyframes feature manual annotations of task-relevant entities, including manipulated objects and robot end-effectors.

\paragraph{Baseline implementation details.}
For RoboEngine-T and RoboEngine-G, we use the official
implementation and released model weights.
For GenAug, we replace its fixed oracle masks with masks
obtained using GroundingDINO and SAM2 to process our
demonstration trajectories.
All augmentation methods retain the recorded robot poses
and action labels.

\subsection{Dual-Factor Generalization Evaluation}
\label{supp:dual}

\textbf{Evaluation Setup.}

\begin{table*}[!t]
  \centering
  \caption{\textbf{Quantitative results under the Dual-Factor Variation setting.} 
    We report the average success rates (\%) on both UR-5e and AgileX robots. 
    The evaluation involves 5 unseen background textures combined with 10 task-irrelevant distractors, totaling 100 trials for each task.}
  \resizebox{0.95\textwidth}{!}{
  \begin{tabular}{l|ccccc|c}
    \toprule
    \textbf{Augmentation Method} & UR-PutCornPlot & UR-MoveLemon & UR-StackBowl & UR-StoreCarrot & UR-OpenDrawerCorn & Average\\
    \cmidrule(lr){1-7}
    \texttt{No Aug} & 0.20 & 0.26 & 0.12 & 0.28 & 0.36 & 0.24 \\
    \texttt{RoboEngine-T} & 0.25 & 0.38 & 0.20 & 0.36 & 0.55 & 0.35 \\
    \texttt{RoboEngine-G} & 0.28 & 0.42 & 0.22 & 0.40 & 0.62 & 0.39\\
    \texttt{GenAug} & 0.30 & 0.44 & 0.20 & 0.46 & 0.68 & 0.42 \\
    \textbf{RoboAug} & \textbf{0.36} & \textbf{0.52} & \textbf{0.28} & \textbf{0.58} & \textbf{0.84} & \textbf{0.52}\\
    \midrule
     & AGX-PutCornPlate & AGX-UprightMug & AGX-StackBowl & AGX-OpenPotCorn & AGX-CloseDrawerCorn & \\
    \cmidrule(lr){1-7}
    \texttt{No Aug} & 0.20 & 0.12 & 0.18 & 0.20 & 0.20 & 0.18 \\
    \texttt{RoboEngine-T} & 0.26 & 0.18 & 0.22 & 0.25 & 0.22 & 0.23 \\
    \texttt{RoboEngine-G} & 0.30 & 0.22 & 0.26 & 0.30 & 0.26 & 0.27\\
    \texttt{GenAug} & 0.28 & 0.20 & 0.28 & 0.26 & 0.32 & 0.27\\
    \textbf{RoboAug} & \textbf{0.56} & \textbf{0.36} & \textbf{0.40} & \textbf{0.40} & \textbf{0.56} & \textbf{0.46}\\
   
    \bottomrule
  \end{tabular}
  }
  \label{tab:dual_factor}
\end{table*}

To rigorously assess the robustness of visual policies under complex environmental shifts, we introduce the \textit{Dual-Factor Variation} protocol. 
This setting challenges the agent with a combination of two distinct perturbations: unseen background textures and object clutter. 
Formally, we evaluate the model using five background textures that were not present in the training set. 
For each background, we introduce 10 task-irrelevant distractors placed randomly across the workspace. 
This configuration aims to verify whether the policy can effectively decouple task-essential features from compounded visual distractions. 
To validate the cross-embodiment stability of RoboAug, we conduct these experiments on two distinct robot embodiments, the UR-5e (UR) and the AgileX (AGX). 
For each background-clutter configuration, we perform 20 evaluation trials, resulting in a total of 100 trials, and report the average success rate.

\textbf{Results under Dual-Factor Variation.}
Quantitative results are summarized in Table~\ref{tab:dual_factor}. 
The complexity of this setting poses a substantial hurdle for existing methods. 
The baseline model, ACT w/o Aug, fails to generalize, yielding average success rates of only $0.24$ and $0.18$ on the UR and AGX suites, respectively. 
While state-of-the-art augmentation methods such as RoboEngine-G and GenAug offer moderate improvements by addressing individual visual factors, their performance degrades significantly when facing simultaneous background and object shifts. 

In contrast, our proposed RoboAug exhibits superior robustness across all evaluated benchmarks. 
On the UR-series tasks, RoboAug achieves an average success rate of $0.52$, surpassing the strongest baseline GenAug by
approximately 10 percentage points.
A consistent trend is observed in the AGX-series, where our method attains an average score of $0.46$. 
These results demonstrate that RoboAug effectively synthesizes a diverse training distribution that captures the underlying visual logic, enabling the model to maintain high robustness even under high-variance dual-factor perturbations.


\subsection{Single-Factor Generalization: Robustness to Unseen Backgrounds.} 
\label{supp:background}

\begin{figure*}[!t]
    \centering
    \includegraphics[width=0.95\linewidth]{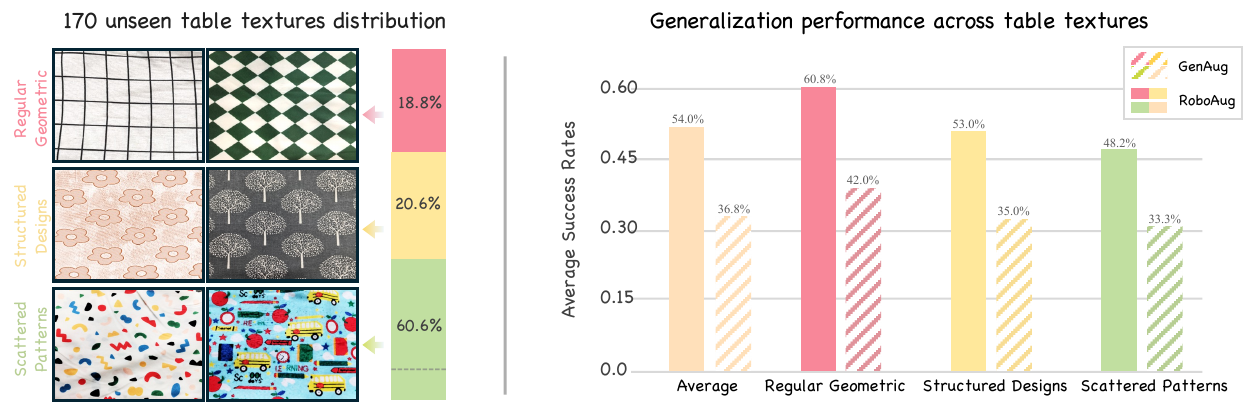}
    \caption{Background generalization on task UR-PutCornPot across 170 unseen backgrounds.}
    \label{fig:background_results}
\end{figure*}

Figure~\ref{fig:background_results} illustrates performance on UR-PutCornPot across the 170 unseen backgrounds. 
RoboAug achieves a mean success rate of 54.0\%, significantly outperforming GenAug (36.8\%). 
While success rates for both methods naturally decrease as background patterns become more intricate, GenAug exhibits a sharper decline. 
These results confirm that RoboAug effectively maintains policy focus on foreground objects despite severe background visual distractions.


\subsection{Single-Factor Generalization: Robustness to Task-Irrelevant Distractors.} 
\label{supp:distractor}

\begin{table*}[t]
  \centering
  \caption{
    Performance Comparison of different methods under a set of 10 distinct distractors.
  }
  \resizebox{0.95\textwidth}{!}{
  \begin{tabular}{l|ccccc|c}
    \toprule
    \textbf{Augmentation Method} & UR-PutCornPlot & UR-MoveLemon & UR-StackBowl & UR-StoreCarrot & UR-OpenDrawerCorn & \textbf{Average}\\
    \cmidrule(lr){1-7}
    \texttt{No Aug} & 0.25 & 0.20 & 0.30 & 0.10 & 0.25 & 0.22\\
    \texttt{RoboEngine-T} & 0.50 & 0.25 & 0.50 & 0.20 & 0.35 & 0.36 \\
    \texttt{RoboEngine-G} & 0.55 & 0.30 & 0.50 & 0.20 & 0.40 & 0.39\\
    \texttt{GenAug} & 0.60 & 0.30 & 0.55 & 0.20 & 0.45 & 0.42\\
    \textbf{RoboAug} & \textbf{0.90} & \textbf{0.45} & \textbf{0.60} & \textbf{0.40} & \textbf{0.50} & \textbf{0.57} \\
    \midrule
     & AGX-PutCornPlate & AGX-UprightMug & AGX-StackBowl & AGX-OpenPotCorn & AGX-CloseDrawerCorn &  \\
    \cmidrule(lr){1-7}
    \texttt{No Aug} & 0.15 & 0.25 & 0.25 & 0.15 & 0.00 & 0.16\\
    \texttt{RoboEngine-T} & 0.20 & 0.30 & 0.40 & 0.30 & 0.15 & 0.27\\
    \texttt{RoboEngine-G} & 0.25 & 0.35 & 0.45 & 0.30 & 0.20 & 0.31\\
    \texttt{GenAug} & 0.30 & 0.35 & 0.55 & 0.25 & 0.10 & 0.31\\
    \textbf{RoboAug} & \textbf{0.45} & \textbf{0.50} & \textbf{0.90} & \textbf{0.40} & \textbf{0.30} & \textbf{0.51}\\
   
    \bottomrule
  \end{tabular}
  }
  \label{tab:distractors_main_results}
\end{table*}

Complementing the background generalization experiments presented in the main manuscript, we further established a single-factor generalization evaluation focused on task-irrelevant distractors.
For each task, we randomly placed 10 task-irrelevant objects on the tabletop and conducted 20 evaluation rollouts.
Table~\ref{tab:distractors_main_results} reports performance under this extreme clutter setting. 
In this challenging scenario, baseline methods exhibit substantial performance degradation. 
Specifically, ACT w/o Aug and RoboEngine-T achieve success rates of only 0.15 and 0.20 on the \texttt{AGX-PutCornPlate} task, respectively, whereas RoboAug attains a markedly higher success rate of 0.45.

Failure inspection indicates that baseline methods often struggle to semantically distinguish the target object from dense background clutter, leading to incorrect object selection or unstable execution.
In contrast, by leveraging the proposed region-contrastive loss, RoboAug learns more discriminative region-level representations, enabling the policy to consistently attend to the task-relevant object and maintain robust performance under heavy visual distraction.

To investigate the impact of clutter density,
Figure~\ref{fig:single-factor} illustrates performance trends as the number of distractors increases from 0 to 10.
On the \texttt{AGX-StackBowl} task, RoboAug consistently outperforms the strongest baseline, GenAug, across all clutter levels.
Notably, RoboAug preserves a high success rate of 0.90 even with 10 distractors, while GenAug experiences a pronounced drop from 0.80 to 0.55.
These results demonstrate that RoboAug is substantially more robust to task-irrelevant visual perturbations

\subsection{Single-Factor Generalization: Robustness to Illumination Variation.} 
\label{supp:light}

\begin{figure}[t]
    \centering
    \includegraphics[width=0.96\linewidth]{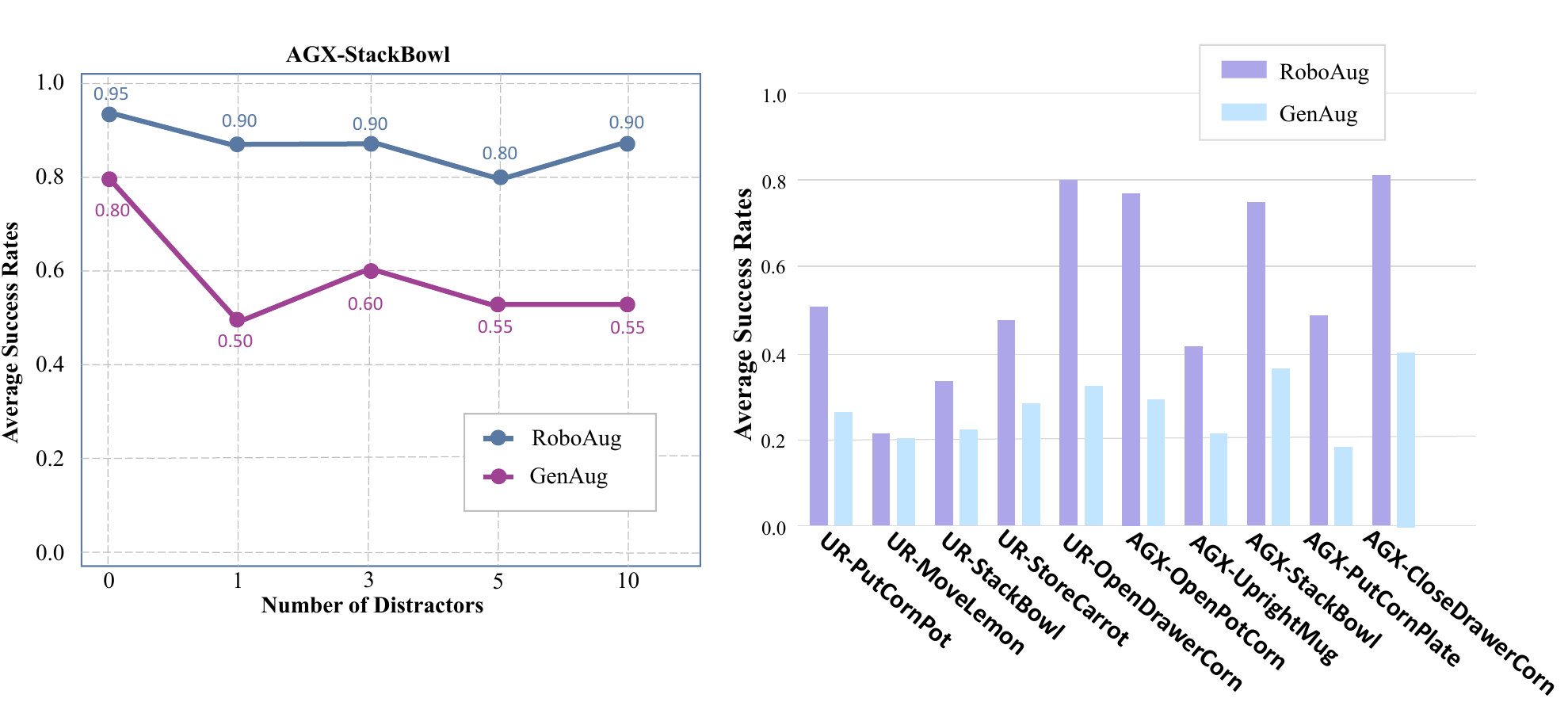}
    \caption{Left: Comparison of task success rates between RoboAug and the best baseline method under varying numbers of distractors; Right: Comparison of RoboAug and best baseline method on 20 lighting conditions, evaluated using the task success rates.}
    \label{fig:single-factor}
\end{figure}


\begin{table*}[t]
  \centering
  \caption{
    Tasks Success Rates on Novel Backgrounds. We evaluated the success rate of augmentation methods across 10 unseen backgrounds.  
  }
  \resizebox{0.95\textwidth}{!}{
  \begin{tabular}{l|ccccc|c}
    \toprule
    \textbf{Method} & UR-PutCornPlot & UR-MoveLemon & UR-StackBowl & UR-StoreCarrot & UR-OpenDrawerCorn & \textbf{Average}\\
    \cmidrule(lr){1-7}
    \texttt{ACT w/o Aug} & 0.24 & 0.15 & 0.12 & 0.22 & 0.34 & 0.21\\
    \texttt{RoboEngine-T} & 0.24 & 0.26 & 0.22 & 0.20 & 0.40 & 0.26 \\
    \texttt{RoboEngine-G} & 0.46 & 0.28 & 0.32 & 0.50 & 0.42 & 0.40 \\
    \texttt{GenAug} & 0.48 & 0.36 & 0.28 & 0.46 & 0.42 & 0.40 \\
    \textbf{RoboAug} & \textbf{0.90} & \textbf{0.60} & \textbf{0.60} & \textbf{0.65} & \textbf{0.65} & \textbf{0.68}\\
    \midrule
     & AGX-PutCornPlate & AGX-UprightMug & AGX-StackBowl & AGX-OpenPotCorn & AGX-CloseDrawerCorn \\
    \cmidrule(lr){1-7}
    \texttt{ACT w/o Aug} & 0.16 & 0.18 & 0.12 & 0.28 & 0.12 & 0.17\\
    \texttt{RoboEngine-T} & 0.22 & 0.20 & 0.38 & 0.20 & 0.30 & 0.26\\
    \texttt{RoboEngine-G} & 0.32 & 0.35 & 0.46 & 0.30 & 0.32 & 0.35\\
    \texttt{GenAug} & 0.36 & 0.30 & 0.47 & 0.40 & 0.42 & 0.39\\
    \textbf{RoboAug} & \textbf{0.74} & \textbf{0.84} & \textbf{0.70} & \textbf{0.52} & \textbf{0.72} & \textbf{0.70}\\
   
    \bottomrule
  \end{tabular}
  }
  \label{tab:add_background}
\end{table*}

As illustrated in Figure~\ref{fig:single-factor}, we further evaluate policy generalization across 10 tasks on the UR-5e and AgileX robots under 20 unseen lighting conditions, including 4 types of dynamic lighting changes.
Under these conditions, the baseline method GenAug exhibits notable performance degradation, particularly in scenarios involving drastic color temperature shifts, strong cast shadows, and dynamic high-contrast illumination.
Such lighting variations alter the apparent shape and texture of objects, frequently leading to perception and recognition failures.

In contrast, RoboAug demonstrates substantially more stable performance across the full range of lighting conditions.
For example, on the \texttt{AGX-OpenPotCorn} task, RoboAug achieves a success rate of 0.75, significantly outperforming GenAug, which attains only 0.30.
These results suggest that background augmentation combined with region-contrastive learning improves robustness to illumination changes, even though lighting variations are not explicitly synthesized during training.

\subsection{Effectiveness of Region-Contrastive Loss across Different Policies}
\label{supp:rcl_policy}

\begin{wrapfigure}{r}{0.50\textwidth}  
    \centering
    \vspace{-20pt}
    \includegraphics[width=\linewidth]{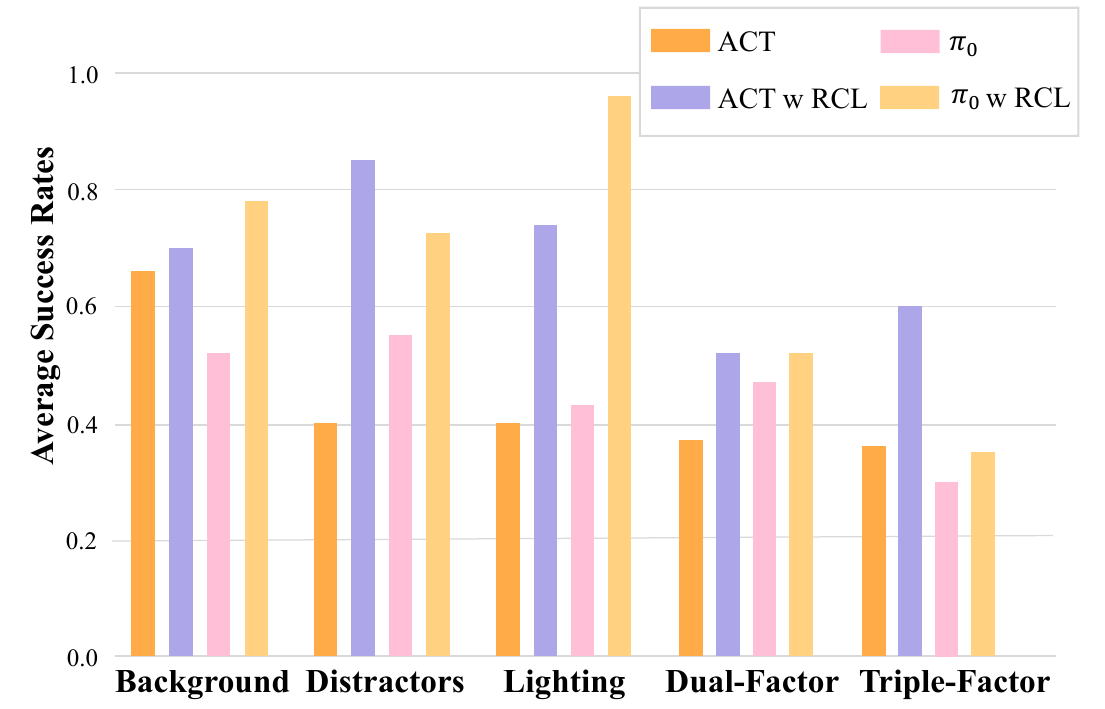} 
    \caption{Effectiveness of RCL.}
    \label{fig:plug_and_play}
     \vspace{-10pt}
\end{wrapfigure}
To verify the universality of our proposed method, we evaluate the impact of the Region-Contrastive Loss (RCL) when integrated into different policy backbones. Specifically, we apply RCL to two representative architectures, ACT~\cite{zhao2023learning} and $\pi_0$~\cite{rss2025_avisionlanguagea}, within the \texttt{AGX-StackBowl} task. 
As illustrated in Figure~\ref{fig:plug_and_play}, the experimental results demonstrate that incorporating RCL consistently improves performance for both policies. 
These findings validate the effectiveness of RCL and suggest that it is compatible with diverse policy formulations.

\subsection{Background Generalization across Multiple Embodiments}
\label{supp:background}

To further validate the effectiveness of RoboAug across different robot configurations and multiple tasks, we conducted extensive experiments focusing on single-factor generalization with respect to background variations. 
Specifically, we evaluated our method on five distinct tasks for both the UR-5e and AgileX robotic arms. 
For each task, we tested the policy on 10 different unseen backgrounds and reported the average success rate. 

Table~\ref{tab:add_background} presents the quantitative results for these experiments. 
As indicated by the data, our proposed method, RoboAug, demonstrates superior generalization capabilities compared to all baselines. 
In the UR-5e robot tasks, RoboAug achieves an average success rate of $68\%$, significantly surpassing the strongest baseline, GenAug, which achieves $40\%$. 
A similar trend is observed in the AgileX robot tasks, where our method reaches a $70\%$ success rate, consistently outperforming other augmentation strategies.

\subsection{Results on Simulation}
\label{sup:sim_result}

\begin{figure}[!t]
    \centering
    \includegraphics[height=0.22\linewidth]{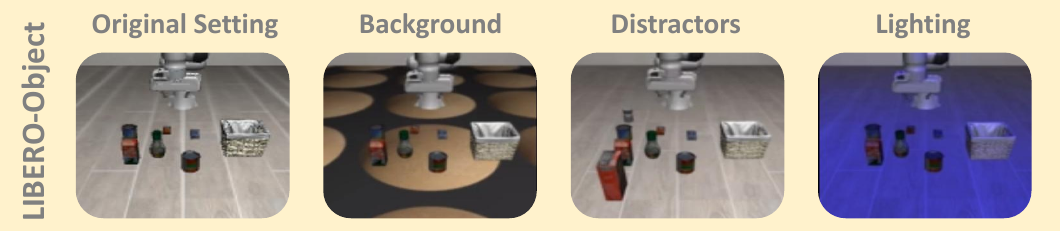}
    \caption{Experimental setup for evaluating generalization on the LIBERO-Plus benchmark.}
    \label{fig:sim_setup}
\end{figure}


\begin{wraptable}{r}{0.48\textwidth}   
    \centering
    \vspace{-10pt}
    \caption{Generalization performance on the LIBERO-Plus benchmark.}
    \label{tab:sim_result}
    \resizebox{\linewidth}{!}{
        \begin{tabular}{c|ccc|c}
            \toprule
            Method & Background & Distractor & Light & Average \\ 
            \midrule
            ACT w/o Aug~\cite{zhao2023learning} & 0.745 & 0.860 & 0.789 & 0.798  \\
            RoboEngine-G~\cite{11247097} & 0.806 & 0.942 & 0.855 & 0.868 \\
            \midrule
            \textbf{RoboAug} & \textbf{0.913} & \textbf{0.990} & \textbf{0.896} & \textbf{0.933} \\
            \bottomrule
        \end{tabular}%
    }
\end{wraptable}
As shown in Figure~\ref{fig:sim_setup}, we utilize the LIBERO-Plus benchmark~\cite{Fei_2026_CVPR}, which introduces diverse environmental shifts to the standard tasks. 
Policies were trained on the LIBERO-Object dataset (10 tasks, 50 demonstrations each) and tested under three perturbation types: background, distractors, and lighting. 
As summarized in Table~\ref{tab:sim_result}, RoboAug consistently outperforms the baselines across all categories. 
Our method achieves an average success rate of 93.3\%, surpassing the strongest baseline, RoboEngine-G, by a significant margin of 6.5\%. 
These results highlight the effectiveness of RoboAug in preventing policy degradation, particularly in scenarios with complex visual distractors and background changes.

\subsection{Additional Ablation Studies}
\label{sec:additional_ablations}

We conduct additional ablation studies on UR-StoreCarrot
and TG2-WeightApple under triple-factor shifts to examine
component contributions, mask quality, reference-frame
sensitivity, and the spatial attention design.
All experiments use ACT with the same training and
evaluation settings, except for the factor being ablated.
Each configuration is evaluated with 20 rollouts per task,
and we report the average success rate across the two tasks.
The original-to-augmented data ratio is 1:5 whenever
background generation is enabled.
Unless otherwise specified, we use the complete RoboAug
configuration with both background generation and RCL.

\paragraph{Contributions of Background Generation and RCL.}
Table~\ref{tab:component_mask_ablation}(a) reports the
contributions of the two components.
Using RCL alone improves the success rate from 0.15
to 0.33, while background generation alone achieves 0.55.
Combining both components further improves performance
to 0.68, indicating that background diversity and
region-contrastive supervision provide complementary
benefits.
Under the matched component setting without RCL,
RoboEngine-G achieves 0.35, compared with 0.55
for our augmentation pipeline.

\paragraph{Effect of Mask Quality.}
We further examine the effect of different mask sources
while keeping background generation and RCL enabled.
As shown in Table~\ref{tab:component_mask_ablation}(b),
detector-only masks achieve a success rate of 0.55.
Our masks improve this to 0.68 using one annotated frame
per task, matching SAM2-only masks that require
50 annotated frames per task.
Oracle masks further improve the success rate to 0.75,
suggesting that more accurate region extraction can
provide additional gains.
These results demonstrate the annotation efficiency
of our pipeline on the two evaluated tasks.

\begin{table}[t]
    \centering
    \caption{
    \textbf{Component and mask-source ablations.}
    (a) Contributions of background generation and RCL,
    using our masks when required.
    (b) Comparison of mask sources with both components
    enabled.
    Ann.\ denotes annotated frames per task;
    SR denotes average success rate.
    Oracle masks use annotations for all frames.
    }
    \label{tab:component_mask_ablation}

    \begin{minipage}[t]{0.47\linewidth}
        \centering
        \small
        \textbf{(a) Component contributions}
        \par\medskip
        \begin{tabular*}{\linewidth}
            {@{\extracolsep{\fill}}cccc@{}}
            \toprule
            Gen.\ Bg. & RCL & Ann. & SR $\uparrow$ \\
            \midrule
            No  & No  & 0 & 0.15 \\
            No  & Yes & 1 & 0.33 \\
            Yes & No  & 1 & 0.55 \\
            Yes & Yes & 1 & \textbf{0.68} \\
            \bottomrule
        \end{tabular*}
    \end{minipage}
    \hfill
    \begin{minipage}[t]{0.49\linewidth}
        \centering
        \small
        \textbf{(b) Mask quality and annotation cost}
        \par\medskip
        \begin{tabular*}{\linewidth}
            {@{\extracolsep{\fill}}lcc@{}}
            \toprule
            Mask source & Ann. & SR $\uparrow$ \\
            \midrule
            Detector-only & 0  & 0.55 \\
            SAM2-only     & 50 & 0.68 \\
            \textbf{Ours} & 1  & 0.68 \\
            \midrule
            Oracle        & All & 0.75 \\
            \bottomrule
        \end{tabular*}
    \end{minipage}
\end{table}

\paragraph{Sensitivity to Reference-Frame Selection.}
We evaluate reference frames with partial occlusion,
unusual lighting, and a rare object pose, and compare
them with a clear reference frame.
As shown in Table~\ref{tab:reference_sensitivity},
the clear reference and the reference with a rare
object pose both achieve a success rate of 0.68.
Unusual lighting reduces performance to 0.65, while
partial occlusion yields 0.60.
Among the tested conditions, partial occlusion has the
largest effect, suggesting that visibility of the
task-relevant regions is an important consideration
when selecting the reference frame.

\begin{table}[t]
    \centering
    \caption{
    \textbf{Sensitivity to reference-frame selection.}
    Average success rates using reference frames with
    different visual conditions, with all other settings
    held fixed.
    }
    \vspace{5pt}
    \label{tab:reference_sensitivity}
    \small
    \begin{tabular*}{0.72\linewidth}
        {@{\extracolsep{\fill}}lc@{}}
        \toprule
        Reference condition & Success rate $\uparrow$ \\
        \midrule
        Clear             & \textbf{0.68} \\
        Partial occlusion & 0.60 \\
        Unusual lighting  & 0.65 \\
        Rare object pose  & \textbf{0.68} \\
        \bottomrule
    \end{tabular*}
\end{table}

\paragraph{Effect of Spatial Attention.}
We compare the proposed attention mechanism with two
variants: computing the attention map from masked-region
features instead of full-image features, and removing
attention gating.
Table~\ref{tab:attention_ablation} shows that attention
computed from full-image features achieves a success
rate of 0.68, whereas both variants achieve 0.15.
These results support the use of full-image features
to guide spatial attention.
A possible explanation is that small masked regions
retain limited spatial information after encoder
downsampling, making it difficult to derive effective
attention weights from masked features alone.

\begin{table}[h]
    \centering
    \caption{
    \textbf{Spatial attention ablation.}
    Average success rates with attention computed from
    full-image features, attention computed from
    masked-region features, and no attention gating.
    All other settings are held fixed.
    }
    \vspace{5pt}
    \label{tab:attention_ablation}
    \small
    \begin{tabular*}{0.85\linewidth}
        {@{\extracolsep{\fill}}lc@{}}
        \toprule
        Attention variant & Success rate $\uparrow$ \\
        \midrule
        Full-image features (ours) & \textbf{0.68} \\
        Masked-region features     & 0.15 \\
        Without attention gating   & 0.15 \\
        \bottomrule
    \end{tabular*}
\end{table}

\subsection{Scaling Law Analysis of Data Augmentation}
\label{sup:scale_law}

\begin{wrapfigure}{r}{0.48\textwidth}   
    \centering
    \vspace{-0.5 cm}
    \includegraphics[width=\linewidth]
    {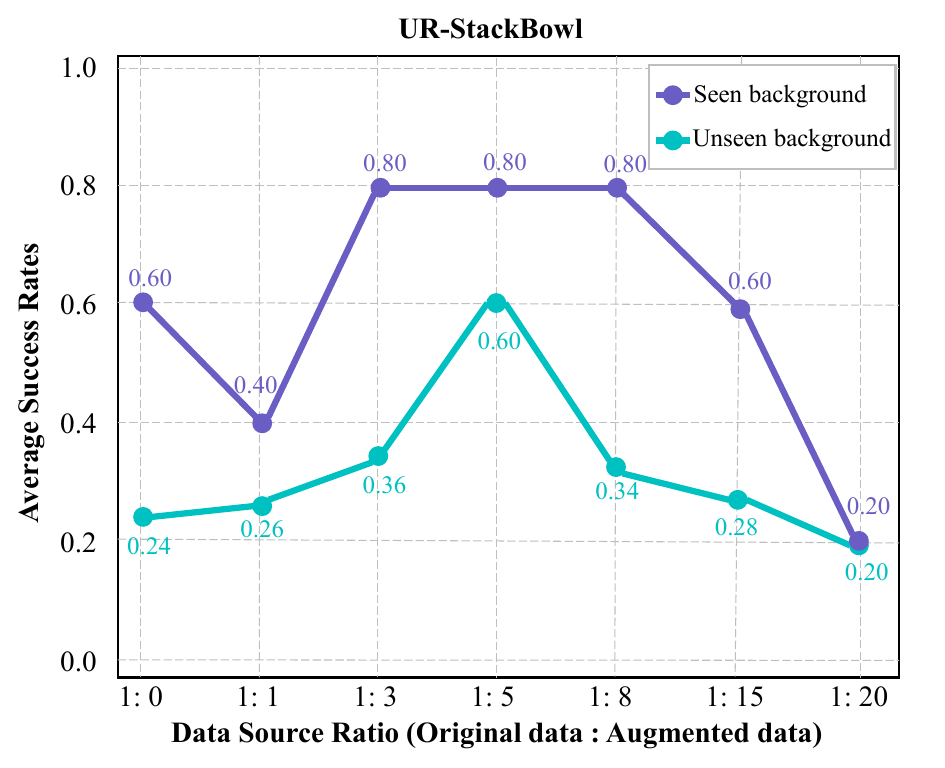} 
    \caption{Effectiveness of data augmentation ratio.}
    \label{fig:scaling_law}
    \vspace{-15pt}
\end{wrapfigure}

To understand the relationship between augmentation magnitude and model performance, we analyze the scaling laws of data augmentation on the UR-StackBowl task by systematically varying the augmentation ratio. 
As shown in Figure~\ref{fig:scaling_law}, 
performance follows a clear inverted U-shaped trend. When the ratio is small, the limited diversity of augmented samples yields only marginal gains over the baseline. 
Moderate settings (1:3 to 1:8) act as effective regularization, injecting sufficient yet controlled variation that prevents overfitting and significantly improves success rates. However, excessive augmentation ($>$ 1:15) overwhelms the network's finite capacity: the introduced visual distribution becomes too far removed from the original task manifold, causing rapid performance deterioration. These findings confirm that a balanced ratio ($\approx$ 1:5), rather than simply maximizing data quantity, is critical for optimal performance.

\subsection{Analysis of Multi-View Data Augmentation }
\label{sup:multi-view}


\begin{wraptable}{r}{0.55\textwidth}   
    \centering
    \vspace{-10pt}
    \caption{Multi-view augmentation performance across three tasks.}
    \label{tab:multi_view}
    \resizebox{\linewidth}{!}{
        \begin{tabular}{lcccc}
            \toprule    
            \textbf{Method} & \textbf{AGX-StackBowl} & \textbf{TG2-CollectBall} & \textbf{TG2-WeighApple} & \textbf{Average} \\
            \midrule
            Single-view w/o RoboAug & 0.14 & 0.16 & 0.20 & 0.18\\
            Single-view w RoboAug & 0.62 & 0.52 & 0.80 & 0.65\\
            Three-view w/o RoboAug & 0.25 & 0.30 & 0.28 & 0.28 \\
            Three-view w RoboAug & 0.70 & 0.64 & 0.85 & 0.73\\
            \bottomrule
        \end{tabular}%
    }
\end{wraptable}
Modern robotic systems increasingly rely on robust multi-view perception, especially for fine-grained manipulation tasks that demand precise hand-eye coordination. To address this, we explore a three-view data augmentation strategy in which the primary view is heavily augmented while two additional views (e.g., the wrist camera) are provided as complementary signals. Our experiments show that combining the augmented primary view with the auxiliary views leads to a marked improvement in task success rates. This suggests that focused augmentation on the principal viewpoint, together with supplementary perspectives, offers a practical and effective way to enhance visual robustness for intricate manipulation.

\subsection{Informal Motivation of RoboAug}
\label{supp:theory}

To provide intuition for the design of RoboAug, we discuss semantic augmentation and region-contrastive learning through the lens of Rademacher complexity. Under idealized assumptions, these two components can be interpreted as increasing visual coverage and
encouraging a more restricted class of policies.
The following analysis provides an informal motivation rather than a formal generalization guarantee for RoboAug. In particular, the independence of augmented samples and exact policy invariance are not established by our method.
\subsubsection{Preliminaries and Definitions}

Let $\mathcal{X}$ and $\mathcal{Y}$ denote the input observation space and action space, respectively. We assume the data is drawn from an underlying distribution $\mathcal{D}$. A policy is a function $\pi: \mathcal{X} \to \mathcal{Y}$ chosen from a hypothesis class $\mathcal{H}$. The goal is to minimize the expected risk $\mathcal{R}(\pi) = \mathbb{E}_{(\mathbf{x}, \mathbf{y}) \sim \mathcal{D}} [\ell(\pi(\mathbf{x}), \mathbf{y})]$, 
where $\ell$ is a bounded continuous loss function.

For this discussion, we decompose an observation $x$ into task-relevant information $x_{\mathrm{task}}$ and task-irrelevant scene factors $x_{\mathrm{scen}}$. We consider label-preserving scene variations for which
the ideal expert action remains unchanged:
$\pi^*(x_{\mathrm{task}}, x_{\mathrm{scen}})
= \pi^*(x_{\mathrm{task}}, x'_{\mathrm{scen}})$.
This is a modeling assumption for the variations
considered here, rather than a claim that arbitrary changes to the environment preserve the correct action.


\subsubsection{Analysis of Semantic Data Augmentation}

Standard generalization bounds depend heavily on the number of training samples $N$. 
We first recall the classical generalization bound based on Rademacher complexity~\cite{bartlett2002rademacher}.

\begin{theorem}[Generalization Bound for Loss Functions]
\label{thm:loss_bound}
Let $\mathcal{H}$ be the policy hypothesis class. Assume the loss function $\ell$ is Lipschitz continuous with respect to its first argument with constant $L_\ell$ and is bounded by $c$. For any $\delta > 0$, with probability at least $1 - \delta$ over the draw of a dataset $S$ of size $N$, the following inequality holds for all $\pi \in \mathcal{H}$:
\begin{equation}
    \mathcal{R}(\pi) \leq \hat{\mathcal{R}}_N(\pi) + 2 L_\ell \mathfrak{R}_{N}(\mathcal{H}) + c\sqrt{\frac{\log(1/\delta)}{2 N}},
\end{equation}
where $\hat{\mathcal{R}}_S(\pi)$ is the empirical risk on the dataset $S$, and $\mathfrak{R}_{N}(\mathcal{H})$ is the Rademacher complexity of $\mathcal{H}$ given $N$ samples.
\end{theorem}

RoboAug expands the original dataset of size $N$
to an augmented dataset of size
$N_{\mathrm{total}} = N + N_{\mathrm{aug}}$.
The resulting images provide additional visual
contexts while retaining the demonstrated action labels.

\paragraph{Idealized Effect of Additional Samples.}
For comparison, consider an idealized dataset
$S_{\mathrm{total}}$ containing $N_{\mathrm{total}}$
independent samples from the same distribution
$\mathcal{D}$.
Under the assumptions of Theorem~6.1, the corresponding
generalization bound takes the form
\begin{equation}
    \mathcal{R}(\pi)
    \leq
    \hat{\mathcal{R}}_{S_{\mathrm{total}}}(\pi)
    + 2L_{\ell}\mathfrak{R}_{N_{\mathrm{total}}}(\mathcal{H})
    + c\sqrt{
        \frac{\log(1/\delta)}{2N_{\mathrm{total}}}
    }.
\end{equation}
For a fixed hypothesis class satisfying
$\mathfrak{R}_{n}(\mathcal{H}) \leq C/\sqrt{n}$,
where $C$ is independent of $n$, the complexity
and concentration terms satisfy
\begin{equation}
    2L_{\ell}\mathfrak{R}_{n}(\mathcal{H})
    + c\sqrt{\frac{\log(1/\delta)}{2n}}
    \leq
    \frac{
        2L_{\ell}C
        + c\sqrt{\log(1/\delta)/2}
    }{\sqrt{n}}.
\end{equation}
This illustrates how additional independent samples
can reduce the estimation terms under the stated
assumptions.

In RoboAug, however, augmented samples share content
and action labels with their source demonstrations.
They are therefore correlated and need not follow
the deployment distribution.
Consequently, $N_{\mathrm{total}}$ cannot generally
be interpreted as the effective number of independent
samples.
The analysis motivates increasing visual diversity,
while the benefit of our augmentation is established
empirically.



\subsubsection{Analysis of Region-Contrastive Learning}
Region-contrastive learning provides a complementary
regularization signal by encouraging consistent
representations of the same task-relevant region
across observations.
To interpret this effect, consider an idealized
hypothesis class
\begin{equation*}
    \mathcal{H}_{\mathrm{inv}}
    =
    \left\{
        \pi \in \mathcal{H}
        \;\middle|\;
        \pi(x_{\mathrm{task}},x_{\mathrm{scen}})
        =
        \pi(x_{\mathrm{task}},x'_{\mathrm{scen}})
        \text{ for admissible scene variations}
    \right\}.
\end{equation*}
This class represents policies that are exactly
invariant to the scene variations under consideration.
RCL encourages representation consistency, but does
not guarantee that the learned policy belongs to
$\mathcal{H}_{\mathrm{inv}}$.

\paragraph{Idealized Effect of Invariance.}
Since $\mathcal{H}_{\mathrm{inv}} \subseteq \mathcal{H}$,
monotonicity of Rademacher complexity gives
\begin{equation}
    \mathfrak{R}_{N_{\mathrm{total}}}
    (\mathcal{H}_{\mathrm{inv}})
    \leq
    \mathfrak{R}_{N_{\mathrm{total}}}(\mathcal{H}).
\end{equation}
The inequality need not be strict.
Under the same independent-sampling assumptions,
a policy $\pi_{\mathrm{inv}} \in
\mathcal{H}_{\mathrm{inv}}$ admits the bound
\begin{equation}
    \mathcal{R}(\pi_{\mathrm{inv}})
    \leq
    \hat{\mathcal{R}}_{S_{\mathrm{total}}}
    (\pi_{\mathrm{inv}})
    + 2L_{\ell}
    \mathfrak{R}_{N_{\mathrm{total}}}
    (\mathcal{H}_{\mathrm{inv}})
    + c\sqrt{
        \frac{\log(1/\delta)}{2N_{\mathrm{total}}}
    }.
\end{equation}
This illustrates the potential benefit of restricting
the policy class when the invariance assumption is
appropriate.
It does not establish a smaller total risk bound
for the policy learned by RoboAug, since neither
exact invariance nor a reduction in empirical risk
is guaranteed.

Taken together, these observations motivate the two
components of RoboAug: semantic augmentation broadens
the visual contexts encountered during training,
while RCL encourages consistent region-level
representations.
The idealized analysis explains why these mechanisms
may be complementary, while their effectiveness
in RoboAug is supported by the experimental results.

\subsection{Instantiations of Generalization Factors}
\label{supp:variant}

In this section, we detail the specific configurations used in our three single-factor generalization experiments, covering background variations, distractor interference, and lighting conditions.

\begin{figure*}[!t]
    \centering
    \includegraphics[width=0.95\linewidth]{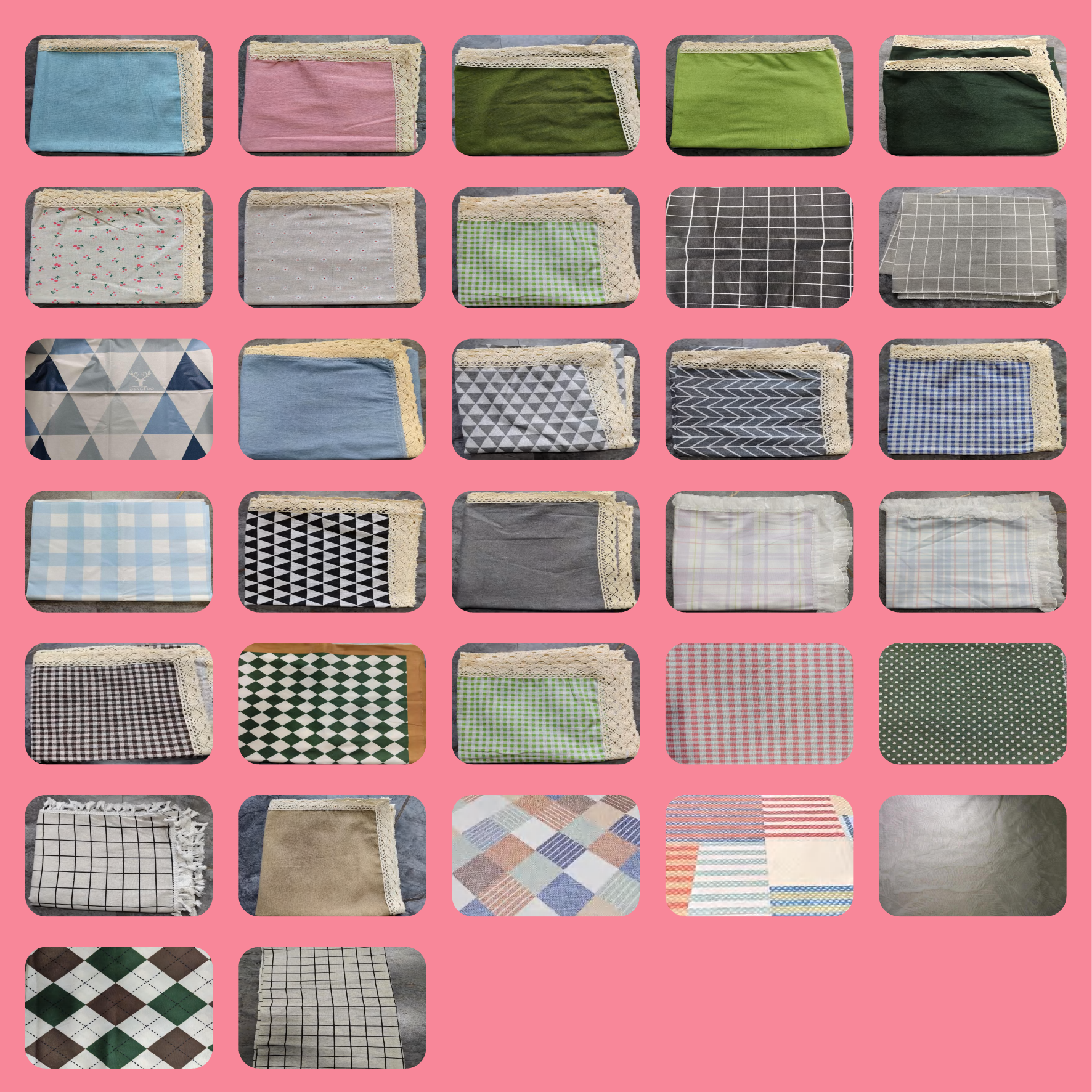}
    \caption{Visualization of the Regular Geometric background category.}
    \label{fig:regular}
\end{figure*}

\begin{figure*}[!t]
    \centering
    \includegraphics[width=0.95\linewidth]{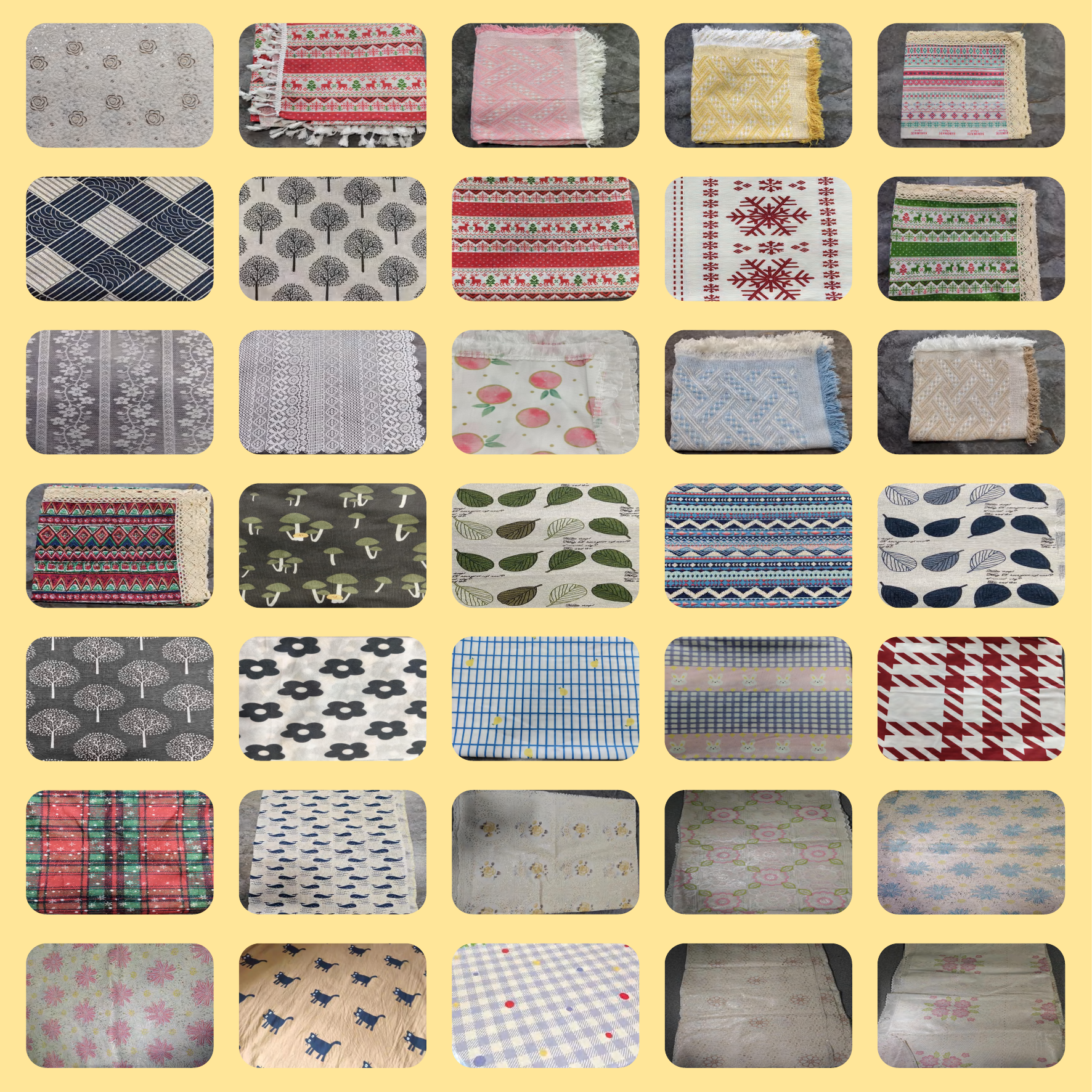}
    \caption{Visualization of the Structured Design background category.}
    \label{fig:structured_design}
\end{figure*}

\begin{figure*}[!t]
    \centering
    \includegraphics[width=0.95\linewidth]{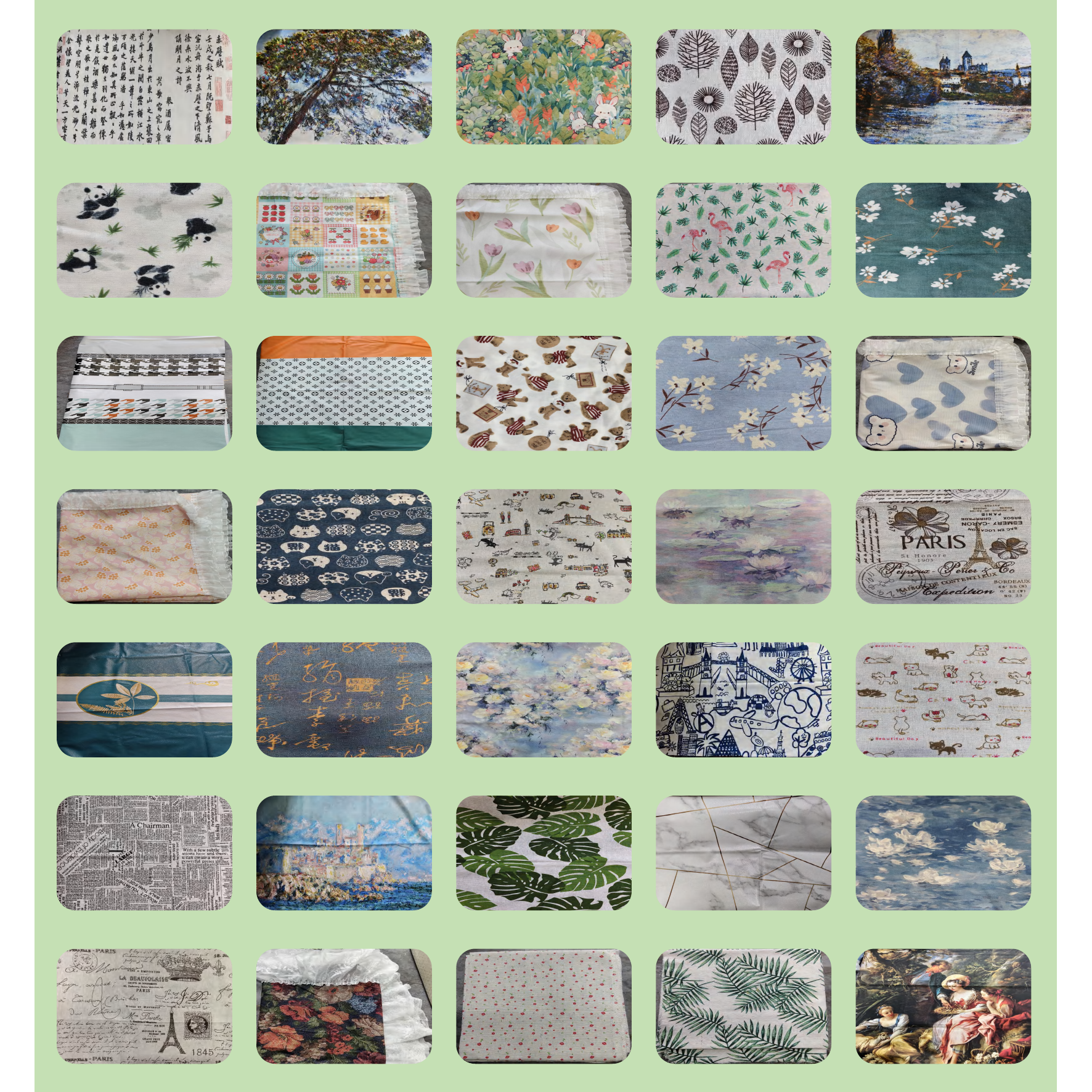}
    \caption{Visualization of the Scattered Pattern background category.}
    \label{fig:scattered_1}
\end{figure*}

\begin{figure*}[!t]
    \centering
    \includegraphics[width=0.95\linewidth]{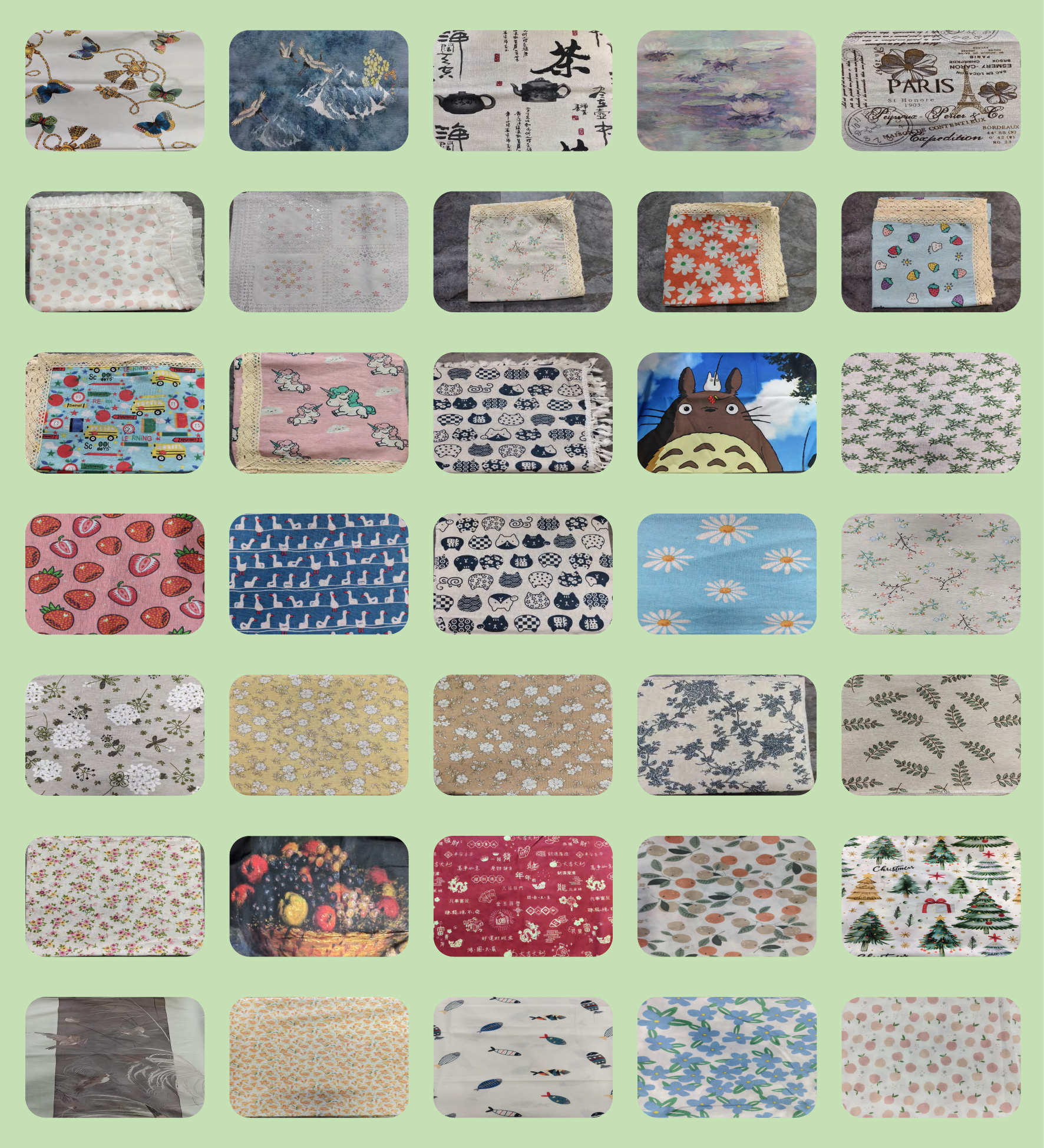}
    \caption{Visualization of the Scattered Pattern background category.}
    \label{fig:scattered_2}
\end{figure*}

\begin{figure*}[!t]
    \centering
    \includegraphics[width=0.95\linewidth]{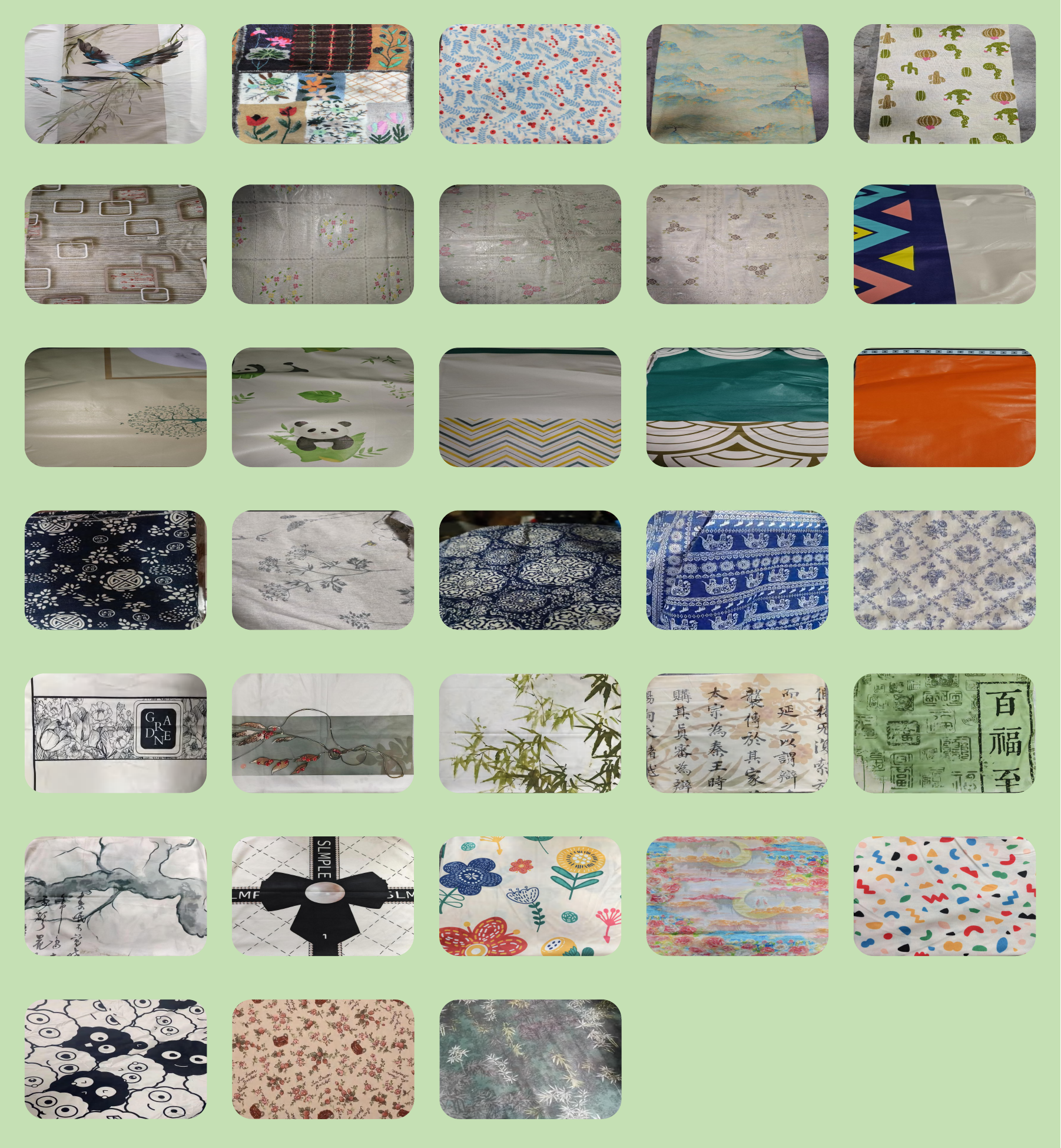}
    \caption{Visualization of the Scattered Pattern background category.}
    \label{fig:scattered_3}
\end{figure*}

\textbf{Background Generalization.}
We evaluated the policy on the UR-CornPot task using 170 distinct, unseen background textures. 
Based on visual complexity, we categorized these backgrounds into three types. 
Visualizations of these 170 background instances are provided in~\Cref{fig:regular,fig:structured_design,fig:scattered_1,fig:scattered_2,fig:scattered_3}.
\begin{itemize}
    \item Regular Geometric: This category comprises 32 patterns characterized by basic shapes and lines, such as squares and rhombuses, arranged in a strictly ordered layout.
    \item Structured Design: This category consists of 35 patterns featuring more intricate motifs, including flowers, leaves, and animals. These patterns maintain a regular, tiled arrangement.
    \item Scattered Pattern: This category includes 103 complex patterns, such as toys and irregular polygons. Unlike the previous categories, these are distributed randomly without a fixed grid, significantly increasing visual interference.
\end{itemize}

\begin{figure*}[!t]
    \centering
    \includegraphics[width=0.85\linewidth]{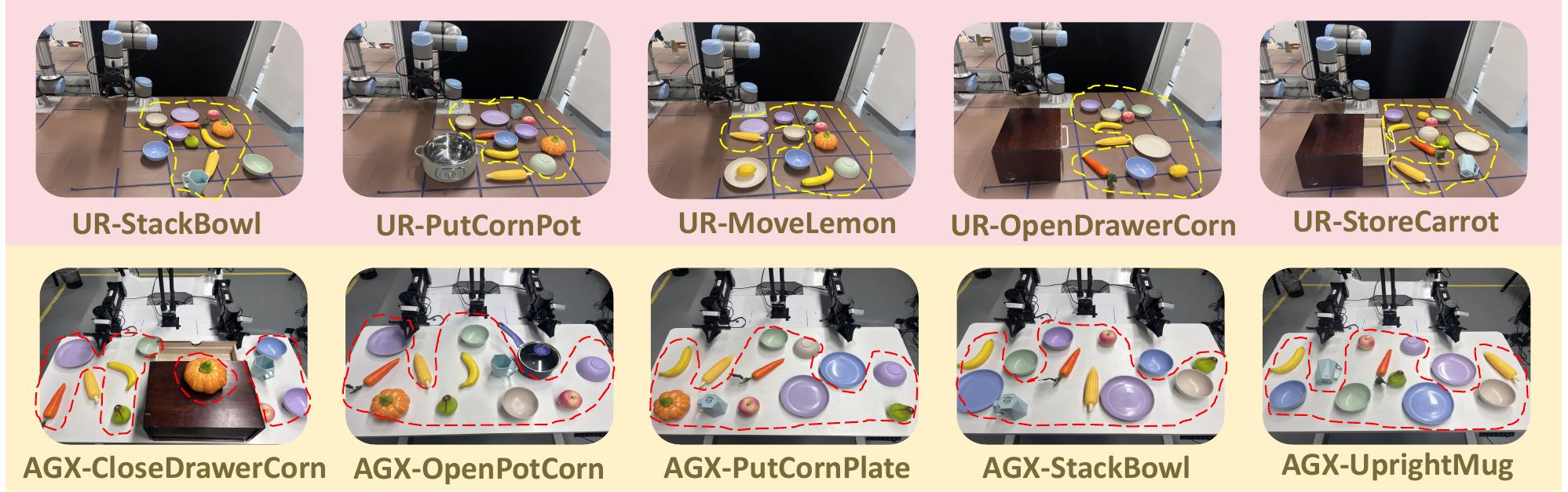}
    \caption{Visualization of the ten distinct distractors used in the UR and AgileX tasks.}
    \label{fig:distractor_variant}
\end{figure*}

\textbf{Distractor Generalization.}
As illustrated in Figure~\ref{fig:distractor_variant}, we introduced visual clutter to test the model's robustness against obstacles. 
We placed up to 10 distractor objects on the tabletop, effectively occupying the entire workspace. 
This setup creates a highly cluttered environment that poses a substantial challenge to the policy.

\begin{figure*}[!t]
    \centering
    \includegraphics[width=0.85\linewidth]{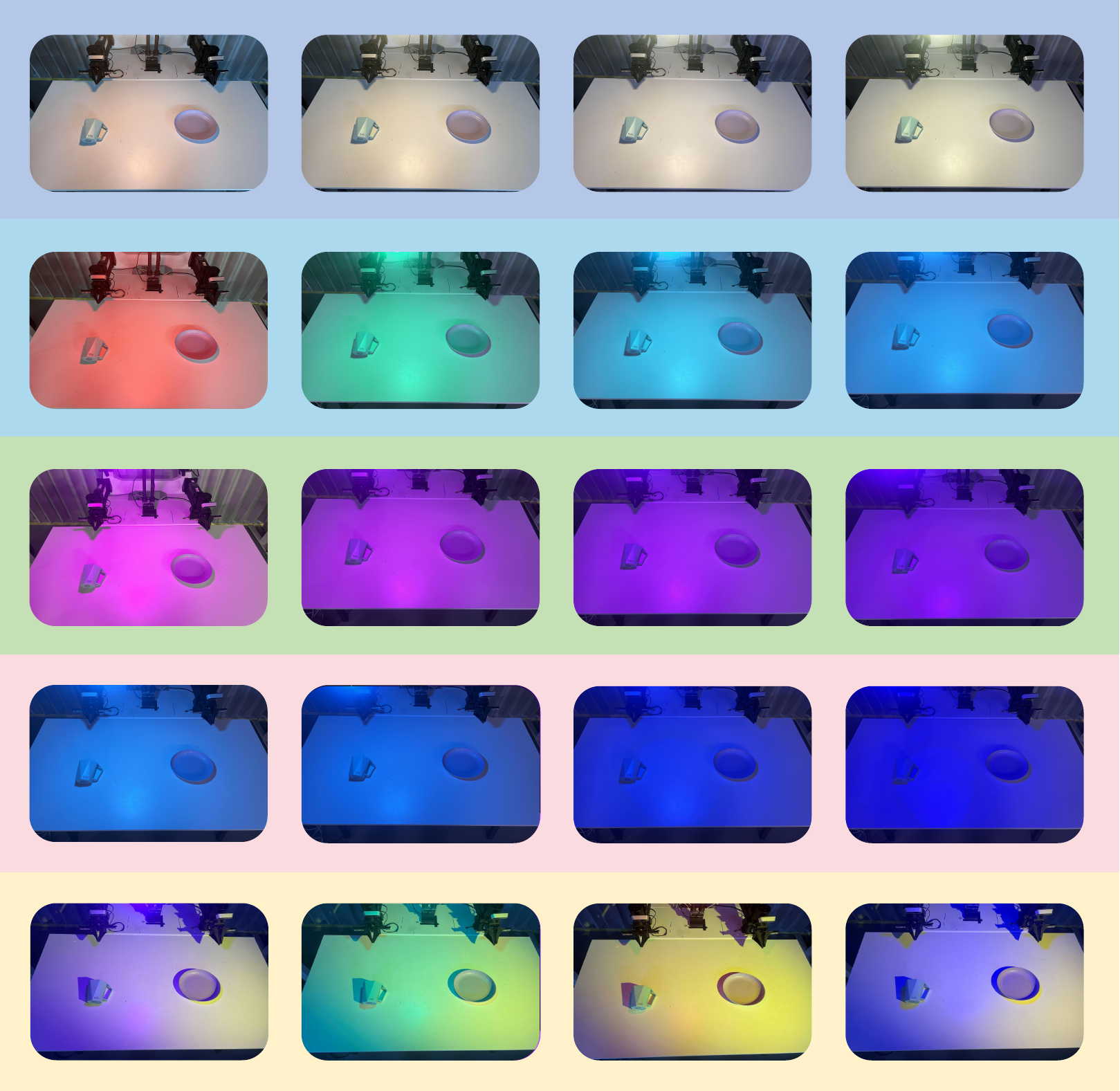}
    \caption{Visualization of twenty distinct illumination conditions. The bottom row demonstrates dynamic lighting scenarios with multicolor changes at varying speeds.}
    \label{fig:light_variant}
\end{figure*}

\textbf{Lighting Generalization.}
As shown in Figure~\ref{fig:light_variant}, we assessed the model's performance under 20 different illumination settings. 
This set includes diverse static conditions as well as three scenarios involving dynamic lighting changes to evaluate adaptability.

\clearpage

\clearpage

\clearpage

\clearpage

\clearpage

\end{document}